# Convolutional Neural Networks for Classification of Alzheimer's Disease: Overview and Reproducible Evaluation

Junhao Wen[a,b,c,d,e*], Elina Thibeau-Sutre[a,b,c,d,e*], Mauricio Diaz-Melo[e,a,b,c,d], Jorge Samper-González[e,a,b,c,d], Alexandre Routier[e,a,b,c,d], Simona Bottani[e,a,b,c,d], Didier Dormont[e,a,b,c,d,f], Stanley Durrleman[e,a,b,c,d], Ninon Burgos[a,b,c,d,e], Olivier Colliot[a,b,c,d,e,f,g,†] , for the Alzheimer's Disease Neuroimaging Initiative[1] and the Australian Imaging Biomarkers and Lifestyle flagship study of ageing[2]

[a]*Institut du Cerveau et de la Moelle épinière, ICM, F-75013, Paris, France*

[b]*Sorbonne Université, F-75013, Paris, France*

[c]*Inserm, U 1127, F-75013, Paris, France*

[d]*CNRS, UMR 7225, F-75013, Paris, France*

[e]*Inria, Aramis project-team, F-75013, Paris, France*

[f]*AP-HP, Hôpital de la Pitié Salpêtrière, Department of Neuroradiology, F-75013, Paris, France*

[g]*AP-HP, Hôpital de la Pitié Salpêtrière, Department of Neurology, F-75013, Paris, France*

*denotes shared first authorship

[†]Corresponding author:

Olivier Colliot, PhD - olivier.colliot@upmc.fr
ICM – Brain and Spinal Cord Institute
ARAMIS team
Pitié-Salpêtrière Hospital
47-83, boulevard de l'Hôpital, 75651 Paris Cedex 13, France

[1] Data used in preparation of this article were obtained from the Alzheimer's Disease Neuroimaging Initiative (ADNI) database (adni.loni.usc.edu). As such, the investigators within the ADNI contributed to the design and implementation of ADNI and/or provided data but did not participate in analysis or writing of this report. A complete listing of ADNI investigators can be found at: http://adni.loni.usc.edu/wp-content/uploads/how_to_apply/ADNI_Acknowledgement_List.pdf

[2] Data used in the preparation of this article was obtained from the Australian Imaging Biomarkers and Lifestyle flagship study of ageing (AIBL) funded by the Commonwealth Scientific and Industrial Research Organisation (CSIRO) which was made available at the ADNI database (www.loni.usc.edu/ADNI). The AIBL researchers contributed data but did not participate in analysis or writing of this report. AIBL researchers are listed at www.aibl.csiro.au.

# Highlights

- We reviewed the state-of-the-art on classification of AD based on CNN and T1 MRI.
- We unveiled data leakage, leading to biased results, in some reviewed studies.
- We proposed a framework for reproducible evaluation of AD classification methods.
- We demonstrated the use of the proposed framework on three public datasets.
- We assessed generalizability both within a dataset and between datasets.

# Graphical abstract

**Preprocessed data**

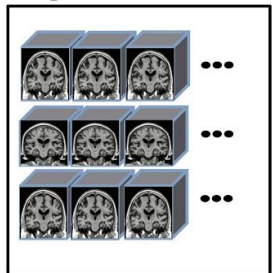

*Data split*

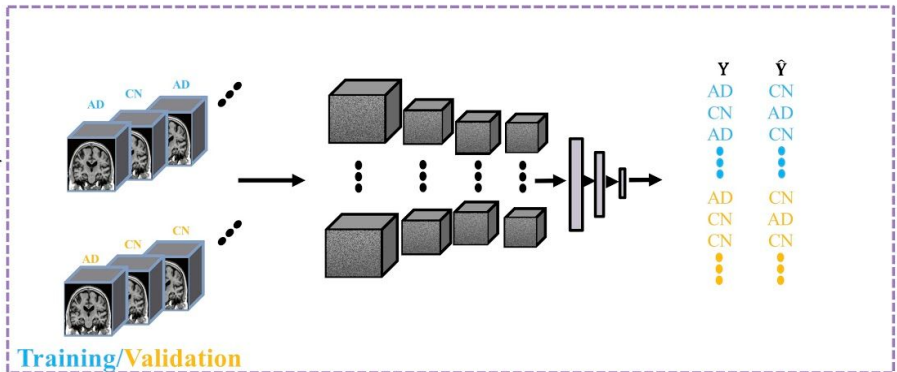

*Image preprocessing*

**BIDS**

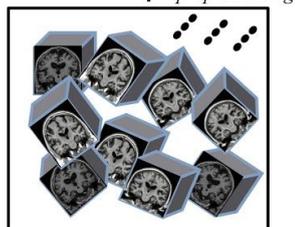

*Automatic converter*

**Downloading**

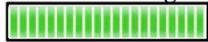

100%

*Model selection & generalization*

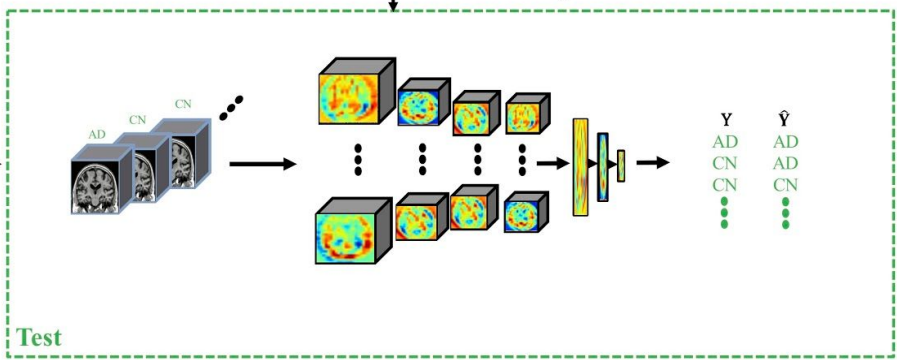

# Abstract

Numerous machine learning (ML) approaches have been proposed for automatic classification of Alzheimer's disease (AD) from brain imaging data. In particular, over 30 papers have proposed to use convolutional neural networks (CNN) for AD classification from anatomical MRI. However, the classification performance is difficult to compare across studies due to variations in components such as participant selection, image preprocessing or validation procedure. Moreover, these studies are hardly reproducible because their frameworks are not publicly accessible and because implementation details are lacking. Lastly, some of these papers may report a biased performance due to inadequate or unclear validation or model selection procedures. In the present work, we aim to address these limitations through three main contributions. First, we performed a systematic literature review. We identified four main types of approaches: i) 2D slice-level, ii) 3D patch-level, iii) ROI-based and iv) 3D subject-level CNN. Moreover, we found that more than half of the surveyed papers may have suffered from data leakage and thus reported biased performance. Our second contribution is the extension of our open-source framework for classification of AD using CNN and T1-weighted MRI. The framework comprises previously developed tools to automatically convert ADNI, AIBL and OASIS data into the BIDS standard, and a modular set of image preprocessing procedures, classification architectures and evaluation procedures dedicated to deep learning. Finally, we used this framework to rigorously compare different CNN architectures. The data was split into training/validation/test sets at the very beginning and only the training/validation sets were used for model selection. To avoid any overfitting, the test sets were left untouched until the end of the peer-review process. Overall, the different 3D approaches (3D-subject, 3D-ROI, 3D-patch) achieved similar performances while that of the 2D slice approach was lower. Of note, the different CNN approaches did not perform better than a SVM with voxel-based features. The different approaches generalized well to similar populations but not to datasets with different inclusion criteria or demographical characteristics. All the code of the framework and the experiments is publicly available: general-purpose tools have been integrated into the Clinica software (www.clinica.run) and the paper-specific code is available at: https://github.com/aramis-lab/AD-DL.

# 1. Introduction

Alzheimer's disease (AD), a chronic neurodegenerative disease causing the death of nerve cells and tissue loss throughout the brain, usually starts slowly and worsens over time (McKhann et al., 1984). AD is expected to affect 1 out of 85 people in the world by the year 2050 (Brookmeyer et al., 2007). The cost of caring for AD patients is also expected to rise dramatically, thus the need of individual computer-aided systems for early and accurate AD diagnosis.

Magnetic resonance imaging (MRI) offers the possibility to study pathological brain changes associated with AD in vivo (Ewers et al., 2011). Over the past decades, neuroimaging data have been increasingly used to characterize AD by means of machine learning (ML) methods, offering promising tools for individualized diagnosis and prognosis (Falahati et al., 2014; Haller et al., 2011; Rathore et al., 2017). A large number of studies have proposed to use predefined features (including regional and voxel-based measurements) obtained from image preprocessing pipelines in combination with different types of classifiers, such as support vector machines (SVM) or random forests. Such approach is often referred to as conventional ML (LeCun et al., 2015). More recently, deep learning (DL), as a newly emerging ML methodology, has made a big leap in the domain of medical imaging (Bernal et al., 2018; J. Liu et al., 2018; Lundervold and Lundervold, 2018; Razzak et al., 2018; D. Wen et al., 2018). As the most widely used architecture of DL, convolutional neural network (CNN) has attracted huge attention due to its great success in image classification (Krizhevsky et al., 2012). Contrary to conventional ML, DL allows the automatic abstraction of low-to-high level latent feature representations (e.g. lines, dots or edges for low level features, and objects or larger shapes for high level features). Thus, one can hypothesize that DL depends less on image preprocessing and requires less prior on other complex procedures, such as feature selection, resulting in a more objective and less bias-prone process (LeCun et al., 2015).

Very recently, numerous studies have proposed to assist diagnosis of AD by means of CNNs (Aderghal et al., 2018, 2017a, 2017b; Bäckström et al., 2018; Basaia et al., 2019; Cheng et al., 2017; Cheng and Liu, 2017; Farooq et al., 2017; Gunawardena et al., 2017; Hon and Khan, 2017; Hosseini Asl et al., 2018; Islam and Zhang, 2018, 2017; Korolev et al., 2017; Lian et al., 2018; Li et al., 2018, 2017; Lin et al., 2018; Manhua Liu et al., 2018; Mingxia Liu et al., 2018a, 2018c; Qiu et al., 2018; Senanayake et al., 2018; Shmulev et al., 2018; Taqi et al., 2018; Valliani and Soni, 2017; Vu et al., 2018, 2017; Wang et al., 2019, 2017; S.-H. Wang et al., 2018; Wu et al., 2018). However, classification results among these studies are not directly comparable because they differ in terms of: i) sets of participants; ii) image preprocessing procedures, iii) cross-validation (CV) procedure and iv) reported evaluation metrics. It is thus impossible to determine which approach performs best. The generalization ability of these approaches also remains unclear. In DL, the use of fully independent test sets is even more critical than in conventional ML, because of the very high flexibility with numerous possible model architecture and training hyperparameter choices. Assessing generalization to other studies is also critical to ensure that the characteristics of the considered study have not been overfitted. In previous works, the generalization may be questionable due to inadequate validation procedures, the absence of an independent test set, or a test set chosen from the same study as the training and validation sets.

In our previous studies (Samper-González et al., 2018; J. Wen et al., 2018), we have proposed an open source framework for reproducible evaluation of AD classification using conventional ML methods. The framework comprises: i) tools to automatically convert three publicly available datasets into the Brain Imaging Data Structure (BIDS) format (Gorgolewski et al., 2016) and ii) a modular set of preprocessing pipelines, feature extraction and classification methods, together with an evaluation framework, that provide a baseline for benchmarking the different components. We demonstrated the use of this framework on positron emission tomography (PET), T1-weighted (T1w) MRI (Samper-González et al., 2018) and diffusion MRI data (J. Wen et al., 2018).

This work presents three main contributions. We first reviewed and summarized the different studies using CNNs and anatomical MRI for AD classification. In particular, we reviewed their validation procedures and the possible presence of data leakage. We then extended our open-source framework for reproducible evaluation of AD classification to DL approaches by implementing a modular set of image preprocessing procedures, classification architectures and evaluation procedures dedicated to DL. Finally, we used this framework to rigorously assess the performance of different CNN architectures, representative of the literature. We studied the influence of key components on the classification accuracy, we compared the proposed CNNs to a conventional ML approach based on a linear SVM, and we assessed the generalization ability of the CNN models within (training and testing on ADNI) and across datasets (training on ADNI and testing on AIBL or OASIS).

All the code of the framework and the experiments is publicly available: general-purpose tools have been integrated into Clinica[3] (Routier et al., 2018), an open-source software platform that we developed to process data from neuroimaging studies, and the paper-specific code is available at: https://github.com/aramis-lab/AD-DL. The tagged version v.0.0.1 corresponds to the version of the code used to obtain the results of the paper. The trained models are available on Zenodo and their associated DOI is 10.5281/zenodo.3491003.

---

## 2. State of the art

We performed an online search of publications concerning classification of AD using neural networks based on anatomical MRI in PubMed and Scopus, from January 1990 to the 15th of January 2019. This resulted in 406 records which were screened according to their abstract, type and content (more details are provided in online supplementary eMethod 1) to retain only those focused on the classification of AD stages using at least anatomical MRI as input of a neural network. This resulted in 71 studies. Out of these 71, 32 studies used CNN on image data in an end-to-end framework, which is the focus of our work.

Depending on the disease stage that is studied, different classification experiments can be performed. We present the main tasks considered in the literature in Section 2.1. We found that a substantial proportion of the studies performed a biased evaluation of results due to the presence of data leakage. These issues are discussed in Section 2.2. We then review the 32 studies that used end-to-end CNNs on image data, the main focus of this work (Section 2.3). Finally, we briefly describe other studies that were kept in our bibliography but that are out of our scope (Section 2.4).

Designing DL approaches for MRI-based classification of AD requires expertise about DL, MRI processing and AD. Such knowledge might be difficult to acquire for newcomers to the field. Therefore, we present a brief introduction to these topics in online supplementary eMethod 2 and 3. Readers can also refer to (Goodfellow et al., 2016) about DL and (Bankman, 2008) for MRI processing.

### 2.1. Main classification tasks

Even though its clinical relevance is limited, differentiating patients with AD from cognitively normal subjects (CN), i.e. AD vs CN, is the most widely addressed task: 25 of the 32 studies presenting an end-to-end CNN framework report results with this task (Table 1). Before the development of dementia, patients go through a phase called mild cognitive impairment (MCI) during which they have objective deficits but not severe enough to result in dementia. Identifying the early stage of AD by differentiating MCI patients from CN subjects (MCI vs CN) is another task of interest, reported in nine studies. Patients with MCI may remain stable or subsequently progress to AD dementia or to another type of dementia. Distinguishing MCI subjects that will progress to AD (denoted as pMCI) from those who will remain stable (denoted as sMCI) would allow predicting the group of subjects that will likely develop the disease. This task (sMCI vs pMCI) has been performed in seven studies. Other experiments performed in the 32 studies on which we focus include differentiating AD from MCI patients (AD vs MCI) and multiclass tasks.

### 2.2. Main causes of data leakage

Unbiased evaluation of classification algorithms is critical to assess their potential clinical value. A major source of bias is data leakage, which refers to the use of test data in any part of the training process (Kriegeskorte et al., 2009; Rathore et al., 2017). Data leakage can be difficult to detect for DL approaches as they can be complex and very flexible. We assessed the prevalence of data leakage among the papers described in section 2.3 and analyzed its causes. The articles were labeled into three categories: i) *Clear* when data leakage was explicitly witnessed; ii) *Unclear* when no sufficient explanation was offered and iii) *None detected*. The results are summarized in the last column of Table 1. They were further categorized according to the cause of data leakage. Four main causes were identified:

1. **Wrong data split.** Not splitting the dataset at the subject-level when defining the training, validation and test sets can result in data from the same subject to appear in several sets. This problem can occur when patches or slices are extracted from a 3D image, or when images of the same subject are available at multiple time points. (Bäckström et al., 2018) showed that, using a longitudinal dataset, a biased dataset split (at the image level) can result in an accuracy increase of 8 percent points compared to an unbiased split (at the subject-level).

2. **Late split.** Procedures such as data augmentation, feature selection or autoencoder (AE) pre-training must never use the test set and thus be performed after the training/validation/test split to avoid biasing the results. For example, if data augmentation is performed before isolating the test data from the training/validation data, then images generated from the same original image may be found in both sets, leading to a problem similar to the wrong data split.

3. **Biased transfer learning.** Transfer learning can result in data leakage when the source and target domains overlap, for example when a network pre-trained on the AD vs CN task is used to initialize a network for the MCI vs CN task and that the CN subjects in the training or validation sets of the source task (AD vs CN) are also in the test set of the target task (MCI vs CN).

4. **Absence of an independent test set.** The test set should only be used to evaluate the final performance of the classifier, not to choose the training hyperparameters (e.g. learning rate) of the model. A separate validation set must be used beforehand for hyperparameter optimization.

Note that we did not consider data leakage occurring when designing the network architecture, possibly chosen thanks to successive evaluations on the test set, as the large majority of the studies does not explicit this step. All these data leakage causes may not have the same impact on data performance. For instance, it is likely that a wrong data split in a longitudinal dataset or at the slice-level is more damaging than a late split for AE pre-training.

## 2.3. Classification of AD with end-to-end CNNs

This section focuses on CNNs applied to an Euclidean space (here a 2D or 3D image) in an end-to-end framework (from the input to the classification). A summary of these studies can be found in Table 1. The table indicates whether data leakage was potentially present, which could have biased the performance upwards. We categorized studies according to the type of input of the network: i) 2D slice-level, ii) 3D patch-level, iii) ROI-based and iv) 3D subject-level.

**Table 1.** Summary of the studies performing classification of AD using CNNs on anatomical MRI. Studies are categorized according to the potential presence of data leakage: (A) studies without data leakage; (B) studies with potential data leakage. The number of citations was found with Google Scholar on 16th of January 2020.

Types of data leakage: a: wrong dataset split; b: absence of independent test set; c: late split; d: biased transfer learning (see Section 2.2).
\* In (Bäckström et al., 2018), data leakage was introduced on purpose to study its influence, which explains its presence in both categories.
† Use of accuracy on a severely imbalanced dataset (one class is less than half of the other), leading to an over-optimistic estimation of performance.
[1] CN vs mild vs moderate vs severe
[2] AD vs MCI vs CN
[3] AD vs LMCI vs EMCI vs CN
[4] sMCI vs pMCI vs CN
ACC: accuracy; BA: balanced accuracy.

(A) None detected Table

| Study | Performance | | | | | Approach | Data leakage | Number of citations |
| | AD vs CN | sMCI vs pMCI | MCI vs CN | AD vs MCI | Multi-class | | | |
| --- | --- | --- | --- | --- | --- | --- | --- | --- |
| (Aderghal et al., 2017b) | ACC=0.84 | -- | ACC=0.65 | ACC=0.67† | -- | ROI-based | None detected | 16 |
| (Aderghal et al., 2018) | BA=0.90 | -- | BA=0.73 | BA=0.83 | -- | ROI-based | None detected | 9 |
| (Bäckström et al., 2018) * | ACC=0.90 | -- | -- | -- | -- | 3D subject-level | None detected | 20 |

| Study | AD vs CN | sMCI vs pMCI | MCI vs CN | AD vs MCI | Multi-class | Approach | Data leakage (type) | Number of citations |
|---|---|---|---|---|---|---|---|---|
| (Cheng et al., 2017) | ACC=0.87 | -- | -- | -- | -- | 3D patch-level | None detected | 12 |
| (Cheng and Liu, 2017) | ACC=0.85 | -- | -- | -- | -- | 3D subject-level | None detected | 8 |
| (Islam and Zhang, 2018) | -- | -- | -- | -- | ACC=0.93[1]† | 2D slice-level | None detected | 23 |
| (Korolev et al., 2017) | ACC=0.80 | -- | -- | -- | -- | 3D subject-level | None detected | 72 |
| (Li et al., 2017) | ACC=0.88 | -- | -- | -- | -- | 3D subject-level | None detected | 12 |
| (Li et al., 2018) | ACC=0.90 | -- | ACC=0.74† | -- | -- | 3D patch-level | None detected | 7 |
| (Lian et al., 2018) | ACC=0.90 | ACC=0.80† | -- | -- | -- | 3D patch-level | None detected | 30 |
| (Mingxia Liu et al., 2018a) | ACC=0.91 | ACC=0.78† | -- | -- | -- | 3D patch-level | None detected | 59 |
| (Mingxia Liu et al., 2018c) | ACC=0.91 | -- | -- | -- | -- | 3D patch-level | None detected | 26 |
| (Qiu et al., 2018) | -- | -- | ACC=0.83† | -- | -- | 2D slice-level | None detected | 8 |
| (Senanayake et al., 2018) | ACC=0.76 | -- | ACC=0.75 | ACC=0.76 | -- | 3D subject-level | None detected | 3 |
| (Shmulev et al., 2018) | -- | ACC=0.62 | -- | -- | -- | 3D subject-level | None detected | 5 |
| (Valliani and Soni, 2017) | ACC=0.81 | -- | -- | -- | ACC=0.57[2] | 2D slice-level | None detected | 8 |

## (B) Data leakage Table

| Study | Performance | | | | | Approach | Data leakage (type) | Number of citations |
|---|---|---|---|---|---|---|---|---|
| | AD vs CN | sMCI vs pMCI | MCI vs CN | AD vs MCI | Multi-class | | | |
| (Aderghal et al., 2017a) | ACC=0.91 | -- | ACC=0.66 | ACC=0.70 | -- | ROI-based | Unclear (b,c) | 13 |
| (Basaia et al., 2019) | BA=0.99 | BA=0.75 | -- | -- | -- | 3D subject-level | Unclear (b) | 25 |
| (Hon and Khan, 2017) | ACC=0.96 | -- | -- | -- | -- | 2D slice-level | Unclear (a,c) | 32 |
| (Hosseini Asl et al., 2018) | ACC=0.99 | -- | ACC=0.94 | ACC=1.00 | ACC=0.95[2] | 3D subject-level | Unclear (a) | 107 |
| (Islam and Zhang, 2017) | -- | -- | -- | -- | ACC=0.74[1]† | 2D slice-level | Unclear (b,c) | 23 |
| (Lin et al., 2018) | ACC=0.89 | ACC=0.73 | -- | -- | -- | ROI-based | Unclear (b) | 22 |
| (Manhua Liu et al., 2018) | ACC=0.85 | ACC=0.74 | -- | -- | -- | 3D patch-level | Unclear (d) | 39 |
| (Taqi et al., 2018) | ACC=1.00 | -- | -- | -- | -- | 2D slice-level | Unclear (b) | 16 |
| (Vu et al., 2017) | ACC=0.85 | -- | -- | -- | -- | 3D subject-level | Unclear (a) | 20 |
| (S.-H. Wang et al., 2018) | ACC=0.98 | -- | -- | -- | -- | 2D slice-level | Unclear (b) | 49 |
| (Bäckström et al., 2018)* | ACC=0.99 | -- | -- | -- | -- | 3D subject-level | Clear (a) | 20 |
| (Farooq et al., 2017) | -- | -- | -- | -- | ACC=0.99[3]† | 2D slice-level | Clear (a,c) | 31 |
| (Gunawardena et al., 2017) | -- | -- | -- | -- | ACC=0.96[2] | 3D subject-level | Clear (a,b) | 8 |
| (Vu et al., 2018) | ACC=0.86 | -- | ACC=0.86 | ACC=0.77 | ACC=0.80[2] | 3D subject-level | Clear (a,c) | 8 |
| (Wang et al., 2017) | -- | -- | ACC=0.91 | -- | -- | 2D slice-level | Clear (a,c) | 11 |
| (Wang et al., 2019) | ACC=0.99 | -- | ACC=0.98 | ACC=0.94 | ACC=0.97[2] | 3D subject-level | Clear (b) | 17 |
| (Wu et al., 2018) | -- | -- | -- | -- | 0.95[4]† | 2D slice-level | Clear (a,b) | 7 |

### 2.3.1.    2D slice-level CNN

Several studies used 2D CNNs with input composed of the set of 2D slices extracted from the 3D MRI volume (Farooq et al., 2017; Gunawardena et al., 2017; Hon and Khan, 2017; Islam and Zhang, 2018, 2017; Qiu et al., 2018; Taqi et al., 2018; Valliani and Soni, 2017; Wang et al., 2017; S.-H. Wang et al., 2018; Wu et al., 2018). An advantage of this approach is that existing CNNs which had huge success for natural image classification, e.g. ResNet (He et al., 2016) and VGGNet (Simonyan and Zisserman, 2014), can be easily borrowed and used in a transfer learning fashion. Another advantage is the increased number of training samples as many slices can be extracted from a single 3D image.

In this subsection of the bibliography, we found only one study in which neither data leakage was detected nor biased metrics were used (Valliani and Soni, 2017). They used a single axial slice per subject (taken in the middle of the 3D volume) to compare the ResNet (He et al., 2016) to an original CNN with only one convolutional layer and two fully connected (FC) layers. They studied the impact of both transfer learning, by initializing their networks with models trained on ImageNet, and data augmentation with affine transformations. They conclude that the ResNet architecture is more efficient than their baseline CNN and that pre-training and data augmentation improve the accuracy of the ResNet architecture.

In all other studies, we detected a problem in the evaluation: either data leakage was present (or at least suspected) (Farooq et al., 2017; Gunawardena et al., 2017; Hon and Khan, 2017; Islam and Zhang, 2017; Taqi et al., 2018; Wang et al., 2017; S.-H. Wang et al., 2018; Wu et al., 2018) or an imbalanced metric was computed on a severely imbalanced dataset (one class is less than half of the other) (Islam and Zhang, 2018; Qiu et al., 2018). Theses studies differ in terms of slice selection: i) one study used all slices of a given plane (except the very first and last ones that are not informative) (Farooq et al., 2017); ii) other studies selected several slices using an automatic (Hon and Khan, 2017; Wu et al., 2018) or manual criterion (Qiu et al., 2018); iii) one study used only one slice (S.-H. Wang et al., 2018). Working with several slices implies to fuse the classifications obtained at the slice-level to obtain a classification at the subject-level. Only one study (Qiu et al., 2018) explained how they performed this fusion. Other studies did not implement fusion and reported the slice-level accuracy (Farooq et al., 2017; Gunawardena et al., 2017; Hon and Khan, 2017; Wang et al., 2017; Wu et al., 2018) or it is unclear if the accuracy was computed at the slice- or subject-level (Islam and Zhang, 2018, 2017; Taqi et al., 2018).

The main limitation of the 2D slice-level approach is that MRI is 3-dimensional, whereas the 2D convolutional filters analyze all slices of a subject independently. Moreover, there are many ways to select slices that are used as input (as all of them may not be informative), and slice-level accuracy and subject-level accuracy are often confused.

### 2.3.2.    3D patch-level CNN

To compensate for the absence of 3D information in the 2D slice-level approach, some studies focused on the 3D patch-level classification (see Table 1). In these frameworks, the input is composed of a set of 3D patches extracted from an image. In principle, this could result, as in the 2D slice-level approach, in a larger sample size, since the number of samples would be the number of patches (and not the number of subjects). However, this potential advantage is not used in the surveyed papers because they trained independent CNNs for each patch. Additional advantages of patches are the lower memory usage, which may be useful when one has limited resources, and the lower number of parameters to learn. However, this last advantage is present only when one uses the same network for all patches.

Two studies (Cheng et al., 2017; Manhua Liu et al., 2018) used very large patches. Specifically, they extracted 27 overlapping 3D patches of size 50×41×40 voxels covering the whole volume of the MR image (100×81×80 voxels). They individually trained 27 convolutional networks (one per patch) comprising four convolutional layers and two FC layers. Then, an ensemble CNN was trained to provide a decision at the subject level. This ensemble CNN is partly initialized with the weights of the previously trained CNNs.

(Manhua Liu et al., 2018) used exactly the same architecture as (Cheng et al., 2017) and enriched it with a fusion of PET and MRI inputs. They also gave the results obtained using the MRI modality only, which is the result reported in Table 1.

(Li et al., 2018) used smaller patches (32×32×32). By decreasing the size of the patches, they had to take into account a possible discrepancy between patches taken at the same coordinates for different subjects. To avoid this dissimilarity between subjects without performing a non-linear registration, they clustered their patches using k-means. Then they trained one CNN per cluster, and assembled the features obtained at the cluster-level in a similar way to (Cheng et al., 2017; Manhua Liu et al., 2018).

The following three studies (Lian et al., 2018; Mingxia Liu et al., 2018a, 2018c) used even smaller patches (19×19×19). Only a subset of patches, chosen based on anatomical landmarks, are used. These anatomical landmarks are found in a supervised manner via a group comparison between AD and CN subjects. This method requires a non-linear registration to build the correspondence between voxels of different subjects. Similarly to other studies, in (Mingxia Liu et al., 2018c), one CNN is pre-trained for each patch and the outputs are fused to obtain the diagnosis of a subject. The approach of (Mingxia Liu et al., 2018a) is slightly different as they consider that a patch cannot be labelled with a diagnosis, hence they do not train one CNN per patch individually before ensemble learning, but train the ensemble network from scratch. Finally, (Lian et al., 2018) proposed a weakly-supervised guidance: the loss of the network is based on the final classification scores at the subject-level as well as the intermediate classification done on the patch- and region-level.

There are far less data leakage problems in this section, with only a doubt about the validity of the transfer learning between the AD vs CN and MCI vs CN tasks in (Manhua Liu et al., 2018) because of a lack of explanations. Nevertheless this has no impact on the result of the AD vs CN task for which we did not detect any problem of data leakage.

As for the 2D-slice level approaches, in which a selection of slices must be made, one must choose the size and stride of patches. The choice of these hyperparameters will depend on the MRI preprocessing (e.g. a non-linear registration is likely needed for smaller patches). Nevertheless, note that the impact of these hyperparameters has been studied in the pre-cited studies (which has not been done for the 2D slice-level approaches). The main drawback of these approaches is the complexity of the framework: one network is trained for each patch position and these networks are successively fused and retrained at different levels of representation (region-level, subject-level).

### 2.3.3. ROI-based CNN

3D patch-level methods use the whole MRI by slicing it into smaller inputs. However, most of these patches are not informative as they contain parts of the brain that are not affected by the disease. Methods based on regions of interest (ROI) overcome this issue by focusing on regions which are known to be informative. In this way, the complexity of the framework can be decreased as fewer inputs are used to train the networks. In all the following studies, the ROI chosen was the hippocampus, which is well-known to be affected early in AD (Dickerson et al., 2001; Salvatore et al., 2015; Schuff et al., 2009). Studies differ by the definition of the hippocampal ROI.

(Aderghal et al., 2018, 2017a, 2017b) performed a linear registration and defined a 3D bounding box comprising all the voxels of the hippocampus according to a segmentation with the AAL atlas. These three studies used a "2D+ε approach" with patches made of three neighbouring 2D slices in the hippocampus. As they use only one or three patches per patient, they do not cover the entire region. The first study (Aderghal et al., 2017b) only uses the sagittal view and classifies one patch per patient. The architecture of the CNN is made of two convolutional layers associated with max pooling, and one FC layer. In the second study (Aderghal et al., 2017a), all the views (sagittal, coronal and axial) are used to generate patches. Then, three patches are generated per subject, and three networks are trained for each view and then fused. The last study from the same author (Aderghal et al., 2018) focuses on the transfer learning from anatomical MRI to diffusion MRI, which is out of our scope.

In (Lin et al., 2018) a non-linear registration was performed to obtain a voxel correspondence between the subjects, and the voxels belonging to the hippocampus[4] were identified after a segmentation implemented with MALP-EM (Ledig et al., 2015). 151 patches were extracted per image with sampling positions fixed during the experiments. Each of them was made of the concatenation of three 2D slices along the three possible planes (sagittal, coronal and axial) originated at one voxel belonging to the hippocampus.

The main drawback of this methodology is that it studies only one (or a few) regions while AD alterations span over multiple brain areas. However, it may reduce the risk of overfitting because the inputs are smaller (~3000 voxels in our bibliography) and fewer than in methods allowing patch combinations.

### 2.3.4. 3D subject-level CNN

Recently, with the boost of high-performance computing resources, more studies used a 3D subject-level approach (see Table 1). In this approach, the whole MRI is used at once and the classification is performed at the subject level. The advantage is that the spatial information is fully integrated.

Some studies readapted two classical architectures, ResNet (He et al., 2016) and VGGNet (Simonyan and Zisserman, 2014), to fit the whole MRI (Korolev et al., 2017; Shmulev et al., 2018). In both cases, the classification accuracies obtained with VGGNet and ResNet are equivalent, and their best accuracies are lower than that of other 3D subject-level approaches. Another study (Senanayake et al., 2018) used a set of complex modules from classical architectures such as ResNet and DenseNet (dilated convolutions, dense blocks and residual blocks), also without success.

Other studies defined original architectures (Bäckström et al., 2018; Basaia et al., 2019; Cheng and Liu, 2017; Hosseini Asl et al., 2018; Li et al., 2017; Vu et al., 2018, 2017; Wang et al., 2019). We detected data leakage in all studies except (Bäckström et al., 2018; Cheng and Liu, 2017; Li et al., 2017). (Bäckström et al., 2018; Cheng and Liu, 2017) had a similar approach by training one network from scratch on augmented data. One crucial difference between these two studies is the preprocessing step: (Bäckström et al., 2018) used non-linear registration whereas (Cheng and Liu, 2017) performed no registration. (Li et al., 2017) proposed a more complex framework fusing the results of a CNN and three networks pre-trained with an AE.

For the other studies using original architectures, we suspect data leakage (Basaia et al., 2019; Hosseini Asl et al., 2018; Vu et al., 2018, 2017; Wang et al., 2019), hence their performance cannot be fairly compared to the previous ones. However we noted that (Hosseini Asl et al., 2018; Vu et al., 2018, 2017) studied the impact of pre-training with an AE, and concluded that it improved their results (accuracy increased from 5 to 10 percent points).

In the 3D-subject level approach, the number of samples is small compared to the number of parameters to optimize. Indeed, there is one sample per subject, typically a few hundreds to thousands of subjects in a dataset, thus increasing the risk of overfitting.

### 2.3.5. Conclusion

A high number of these 32 studies presented a biased performance because of data leakage: 10 were labeled as *Unclear* because of lack of explanations, and 6 as *Clear* (we do not count here the study of Backstrom et al (Bäckström et al., 2018) as data leakage was done deliberately to study its impact). This means that about 50% of the surveyed studies could report biased results.

In addition to that problem, most studies are not comparable because the datasets used, subjects selected among them and preprocessing performed are different. Furthermore, these studies often do not motivate the choice of their architecture or hyperparameters. It might be that many of them have been tried (but not reported) thereby resulting in a biased performance on the test set. Finally, the code and key implementation details (such as hyperparameter values) are often not available, making them difficult if not impossible to reproduce.

---

[4] In their original paper, this anatomical structure was called the "hippopotamus" (sic).

## 2.4.    Other deep learning approaches for AD classification

Several studies found during our literature search are out of our scope: either CNNs were not used in an end-to-end manner or not applied to images, other network architectures were implemented, or the approach required longitudinal or multimodal data.

In several studies, the CNN is used as a feature extractor only and the classification is performed using either a random forest (Chaddad et al., 2018), SVM with linear or polynomial kernels and logistic regression (Çitak-ER et al., 2017), extreme ML (Lin et al., 2018), SVM with different kernels (Shen et al., 2018), or logistic regression and XGBoost (decision trees) (Shmulev et al., 2018). Only Shmulev et al. compared the results obtained with the CNN classification with those obtained with other classifiers based on features extracted by the CNN, and concluded that the latter is more efficient. Instead of being directly applied to the image, CNNs can be applied to pre-extracted features. This is the case of (Suk et al., 2017) where the CNN is applied to the outputs of several regression models performed between MRI-based features and clinical scores with different hyperparameters. CNNs can also be applied to non-Euclidean spaces, such as graphs of patients (Parisot et al., 2018) or the cortical surface (Mostapha et al., 2018).

Other architectures have been applied to anatomical MRI. Many studies used a variant of the multilayer perceptron composed of stacked FC layers (Amoroso et al., 2018; Baskar et al., 2018; Cárdenas-Peña et al., 2017, 2016; Dolph et al., 2017; Gorji and Haddadnia, 2015; Gutiérrez-Becker and Wachinger, 2018; Jha et al., 2017; Lu et al., 2018; Mahanand et al., 2012; Maitra and Chatterjee, 2006; Ning et al., 2018; Raut and Dalal, 2017; Shams-Baboli and Ezoji, 2017; Zhang et al., 2018; Zhou et al., 2019) or of a probabilistic neural network (Duraisamy et al., 2019; Mathew et al., 2018). In other studies, high-level representations of the features are extracted using both unsupervised (deep Boltzmann machine (Suk et al., 2014) and AE (Suk et al., 2015)) and supervised structures (deep polynomial networks (Shi et al., 2018)), and an SVM is used for classification. Non-CNN architectures require extensive preprocessing as they have to be applied to imaging features such as cortical thickness, shapes, or texture, and regional features. Moreover, feature selection or embedding is also often required (Amoroso et al., 2018; Dolph et al., 2017; Jha et al., 2017; Lu et al., 2018; Mahanand et al., 2012; Mathew et al., 2018; Suk et al., 2015, 2014) to further reduce dimensionality.

DL-based classification approaches are not limited to cross-sectional anatomical MRI. Longitudinal studies exploit information extracted from several time points of the same subject. A specific structure, the recurrent neural network, has been used to study the temporal correlation between images (Bhagwat et al., 2018; Cui et al., 2018; X. Wang et al., 2018). Several studies exploit multi-modal data (Aderghal et al., 2018; Cheng and Liu, 2017; Esmaeilzadeh et al., 2018; Li et al., 2015; Liu et al., 2016, 2015; Manhua Liu et al., 2018; Mingxia Liu et al., 2018b; Lu et al., 2018; Ning et al., 2018; Ortiz et al., 2016; Qiu et al., 2018; Raut and Dalal, 2017; Senanayake et al., 2018; Shi et al., 2018; Shmulev et al., 2018; Spasov et al., 2018; Suk et al., 2014; Thung et al., 2017; Vu et al., 2018, 2017; Zhou et al., 2019, 2017), such as multiple imaging modalities (PET and diffusion tensor imaging), demographic data, genetics, clinical scores, or cerebrospinal fluid biomarkers. Note that multimodal studies that also reported results with MRI only (Aderghal et al., 2018; Cheng and Liu, 2017; Manhua Liu et al., 2018; Qiu et al., 2018; Senanayake et al., 2018; Shmulev et al., 2018; Vu et al., 2018, 2017) are displayed in Table 1. Exploiting multiple time-points and/or modalities is expected to improve the classification performance. However, these studies can be limited by the small number of subjects having all the required time points and modalities.

## 3.  Materials

The data used in our study are from three public datasets: the Alzheimer's Disease Neuroimaging Initiative (ADNI) study, the Australian Imaging, Biomarkers and Lifestyle (AIBL) study and the Open Access Series of Imaging Studies (OASIS). These datasets are described in supplementary eMethod 4. We used the T1w MRI available in each of these studies. For the detailed MRI protocols, one can see (Samper-González et al., 2018).

The ADNI dataset used in our experiments comprises 1455 participants for whom a T1w MR image was available at at least one visit. Five diagnosis groups were considered:

- CN: sessions of subjects who were diagnosed as CN at baseline and stayed stable during the follow-up;

- AD: sessions of subjects who were diagnosed as AD at baseline and stayed stable during the follow-up;

- MCI: sessions of subjects who were diagnosed as MCI, EMCI or LMCI at baseline, who did not encounter multiple reversions and conversions and who did not convert back to CN;

- pMCI: sessions of subjects who were diagnosed as MCI, EMCI or LMCI at baseline, and progressed to AD during the 36 months following the current visit;

- sMCI: sessions of subjects who were diagnosed as MCI, EMCI or LMCI at baseline, and did not progress to AD during the 36 months following the current visit.

AD and CN subjects whose label changed over time were excluded. This was also the case for MCI patients with two or more label changes (for instance progressing to AD and then reverting back to MCI). We made this choice because one can assume that the diagnosis of these subjects is less reliable. Naturally, all the sessions of the pMCI and sMCI groups are included in the MCI group. Note that the reverse is false, as some MCI subjects did not convert to AD but were not followed long enough to state whether they were sMCI. Moreover, for 30 sessions, the preprocessing did not pass the quality check (QC) (see Section 4.2) and these images were removed from our dataset. Two pMCI subjects were entirely removed because the preprocessing failed for all their sessions. Table 2 summarizes the demographics, and the MMSE and global CDR scores of the ADNI participants.

**Table 2**. Summary of participant demographics, mini-mental state examination (MMSE) and global clinical dementia rating (CDR) scores at baseline for ADNI.

|        | Subjects | Sessions | Age                      | Gender          | MMSE                   | CDR                       |
|--------|----------|----------|--------------------------|-----------------|------------------------|---------------------------|
| CN     | 330      | 1 830    | 74. 4 ± 5.8 [59.8, 89.6] | 160 M / 170 F   | 29.1 ± 1.1 [24, 30]    | 0: 330                    |
| MCI    | 787      | 3 458    | 73.3 ± 7.5 [54.4, 91.4]  | 464 M / 323 F   | 27.5 ± 1.8 [23, 30]    | 0: 2; 0.5: 785            |
| sMCI   | 298      | 1 046    | 72.3 ± 7.4 [55.0, 88.4]  | 175 M / 123 F   | 28.0 ± 1.7 [23, 30]    | 0.5: 298                  |
| pMCI   | 295      | 865      | 73.8 ± 6.9 [55.1, 88.3]  | 176 M / 119 F   | 26.9 ± 1.7 [23, 30]    | 0.5: 293; 1: 2            |
| AD     | 336      | 1 106    | 75.0 ± 7.8 [55.1, 90.9]  | 185 M / 151 F   | 23.2 ± 2.1 [18, 27]    | 0.5: 160; 1: 175; 2: 1    |

*Values are presented as mean ± SD [range]. M: male, F: female*

The AIBL dataset considered in this work is composed of 598 participants for whom a T1w MR image and an age value was available at at least one visit. The criteria used to create the diagnosis groups are identical to the ones used for ADNI. Table 3 summarizes the demographics, and the MMSE and global CDR scores of the AIBL participants. After the preprocessing pipeline, seven sessions were removed without changing the number of subjects.

**Table 3**. Summary of participant demographics, mini-mental state examination (MMSE) and global clinical dementia rating (CDR) scores at baseline for AIBL.

| | N | Age | Gender | MMSE | CDR |
|---|---|---|---|---|---|
| CN | 429 | 72.5 ± 6.2 [60, 92] | 183 M / 246 F | 28.8 ± 1.2 [25, 30] | 0: 406; 0.5: 22; 1: 1 |
| MCI | 93 | 75.4 ± 6.9 [60, 96] | 50 M / 43 F | 27.0 ± 2.1 [20, 30] | 0: 6; 0.5: 86; 1: 1 |
| sMCI | 13 | 76.7 ± 6.5 [64, 87] | 8 M / 5 F | 28.2 ± 1.5 [26, 30] | 0.5: 13 |
| pMCI | 20 | 78.1 ± 6.6 [63, 91] | 10 M / 10 F | 26.7 ± 2.1 [22, 30] | 0.5: 20 |
| AD | 76 | 73.9 ± 8.0 [55, 93] | 33 M / 43 F | 20.6 ± 5.5 [6, 29] | 0.5: 31; 1: 36; 2: 7; 3: 2 |

*Values are presented as mean ± SD [range]. M: male, F: female*

The OASIS dataset considered in this work is composed of 193 participants aged 62 years or more (minimum age of the participants diagnosed with AD). Table 4 summarizes the demographics, and the MMSE and global CDR scores of the OASIS participants. After the preprocessing pipeline, 22 AD and 17 CN subjects were excluded.

**Table 4**. Summary of participant demographics, mini-mental state examination (MMSE) and global clinical dementia rating (CDR) scores for OASIS.

| | N | Age | Gender | MMSE | CDR |
|---|---|---|---|---|---|
| CN | 76 | 76.5 ± 8.4 [62, 94] | 14 M / 62 F | 29.0 ± 1.2 [25, 30] | 0: 76 |
| AD | 78 | 75.6 ± 7.0 [62, 96] | 35 M / 43 F | 24.4 ± 4.3 [14, 30] | 0.5: 56; 1: 20; 2: 2 |

*Values are presented as mean ± SD [range]. M: male, F: female*

Note that for the ADNI and AIBL datasets, three diagnosis labels (CN, MCI and AD) exist and are assigned by a physician after a series of clinical tests (Ellis et al., 2010, 2009; Petersen et al., 2010) while for OASIS only two diagnosis labels exist, CN and AD (the MCI subjects are labelled as AD), and it is assigned based on the CDR only (Marcus et al., 2007). As the diagnostic criteria of these studies differ, there is no strict equivalence between the labels of ADNI and AIBL, and those of OASIS.

## 4.   Methods

In this section, we present the main components of our framework: automatic converters of public datasets for reproducible data management (Section 4.1), preprocessing of MRI data (4.2), classification models (4.3), transfer learning approaches (4.4), classification tasks (4.5), evaluation strategy (4.6) and framework implementation details (4.7).

### 4.1.   Converting datasets to a standardized data structure

ADNI, AIBL and OASIS, as public datasets, are extremely useful to the research community. However, they may be difficult to use because the downloaded raw data do not possess a clear and uniform organization. We thus used our previously developed converters (Samper-González et al., 2018) (available in the open source software platform Clinica) to convert the raw data into the BIDS format (Gorgolewski et al., 2016). Finally, we organized all the outputs of the experiments into a standardized structure, inspired from BIDS.

### 4.2.   Preprocessing of T1w MRI

In principle, CNNs require only minimal preprocessing because of their ability to automatically extract low-to-high level features. However, in AD classification where datasets are relatively small and thus deep networks may be difficult to train, it remains unclear whether they can benefit from more extensive preprocessing. Moreover, previous studies have used varied preprocessing procedures but without systematically assessing their impact. Thus, in the current study, we compared two different image preprocessing procedures: a "Minimal" and a more "Extensive" procedure. Both procedures included bias field correction, and (optional) intensity rescaling. In addition, the "Minimal" processing included a linear registration while the "Extensive" included non-linear registration and skull-stripping. The essential MR image processing steps to consider in the context of AD classification are presented in online supplementary eMethod 3.

In brief, the "Minimal" preprocessing procedure performs the following operations. The N4ITK method (Tustison et al., 2010) was used for bias field correction. Next, a linear (affine) registration was performed using the SyN algorithm from ANTs (Avants et al., 2008) to register each image to the MNI space (ICBM 2009c nonlinear symmetric template) (Fonov et al., 2011, 2009). To improve the computational efficiency, the registered images were further cropped to remove the background. The final image size is $169 \times 208 \times 179$ with 1 $mm^3$ isotropic voxels. Intensity rescaling, which was performed based on the min and max values, denoted as MinMax, was set to be optional to study its influence on the classification results.

In the "Extensive" preprocessing procedure, bias field correction and non-linear registration were performed using the Unified Segmentation approach (Ashburner and Friston, 2005) available in SPM12[5]. Note that we do not use the tissue probability maps but only the nonlinearly registered, bias corrected, MR images. Subsequently, we perform skull-stripping based on a brain mask drawn in MNI space. We chose this mask-based approach over direct image-based skull-stripping procedures because the later did not prove robust on our data. This mask-based approach is less accurate but more robust. In addition, we performed intensity rescaling as in the "Minimal" pipeline.

We performed QC on the outputs of the preprocessing procedures. For the "Minimal" procedure, we used a DL-based QC framework[6] (Fonov et al., 2018) to automatically check the quality of the linearly registered data. This software outputs a probability indicating how accurate the registration is. We excluded the scans with a probability lower than 0.5 and visually checked the remaining scans whose probability were lower than 0.70. As a result, 30 ADNI scans, 7 AIBL scans, and 39 OASIS scans were excluded.

---

[5] http://www.fil.ion.ucl.ac.uk/spm/software/spm12/
[6] https://github.com/vfonov/deep-qc

### 4.3. Classification models

We considered four different classification approaches: i) 3D subject-level CNN, ii) 3D ROI-based CNN, iii) 3D patch-level CNN and iv) 2D slice-level CNN.

In the case of DL, one challenge is to find the "optimal" model (i.e. global minima), including the architecture hyperparameters (e.g. number of layers, dropout, batch normalization) and the training hyperparameters (e.g. learning rate, weight decay). We first reviewed the architectures used in the literature among the studies in which no data leakage problem was found (Table 1A). As there was no consensus, we used the following heuristic strategy for each of the four approaches.

For the 3D subject-level approach, we began with an overfitting model that was very heavy because of the high number of FC layers (4 convolutional blocks + 5 FC layers). Then, we iteratively repeated the following operations:

- the number of FC layers was decreased until accuracy on the validation set decreased substantially;
- we added one more convolutional block.

In this way, we explored the architecture space from 4 convolutional blocks + 5 FC layers to 7 convolutional blocks + 2 FC layers. Among the best performing architectures, we chose the shallowest one: 5 convolutional blocks + 3 FC layers.

As the performance was very similar for the different architectures tested with the 3D subject-level approach, and as this search method is time costly, it was not used for the 3D patch-level approach for which only four different architectures were tested:

- 4 convolutional blocks + 2 FC layers
- 4 convolutional blocks + 1 FC layer
- 7 convolutional blocks + 2 FC layers
- 7 convolutional blocks + 1 FC layer

The best architecture (4 convolutional blocks + 2 FC layers) was used for both the 3D patch-level and ROI-based approaches. Note that the other architectures were only slightly worse.

For these 3 approaches, other architecture hyperparameters were explored: with or without batch normalization, with or without dropout.

For the 2D slice-level approach, we chose to use a classical architecture, the ResNet-18 with FC layers added at the end of the network. We explored from 1 to 3 added FC layers and the best results were obtained with one. We then explored the number of layers to fine-tune (2 FC layers or the last residual block + 2 FC layers) and chose to fine-tune the last block and the 2 FC layers. We always used dropout and tried different dropout rates.

For all four approaches, training hyperparameters (learning rate, weight decay) were adapted for each model depending on the evolution of the training accuracy.

The list of the chosen architecture hyperparameters is given in online supplementary eTables 1, 2 and 3. The list of the chosen training hyperparameters is given in online supplementary eTables 4 and 5.

#### 4.3.1. 3D subject-level CNN

For the 3D-subject-level approach, the proposed CNN architecture is shown in Figure 1. The CNN consisted of 5 convolutional blocks and 3 FC layers. Each convolutional block was sequentially made of one convolutional layer, one batch normalization layer, one ReLU and one max pooling layer (more architecture details are provided in online supplementary eTable 1).

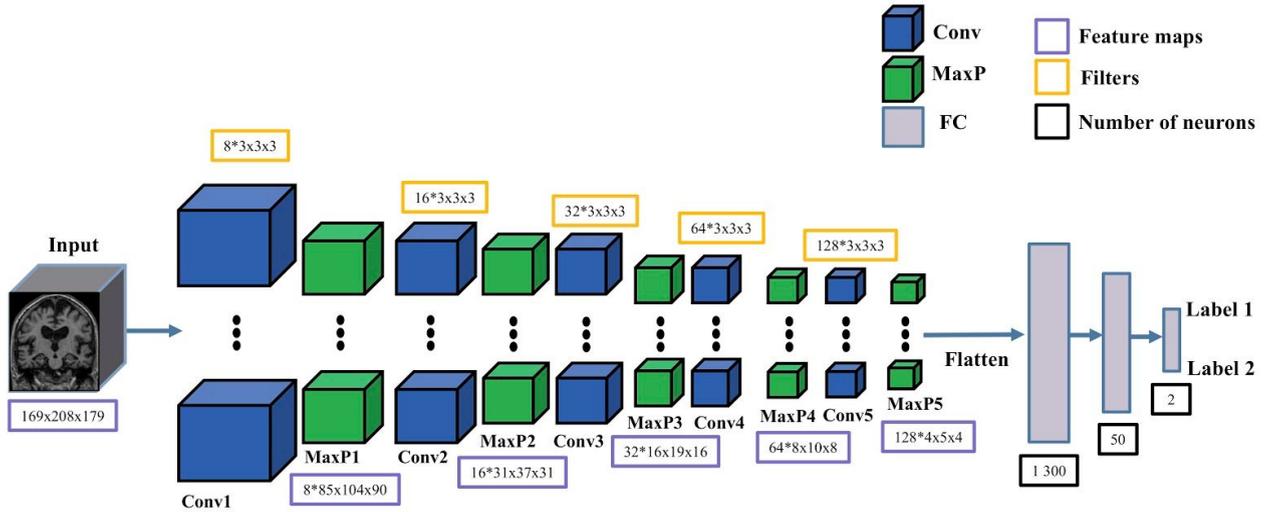

**Figure 1**: Architecture of the 3D subject-level CNNs. For each convolutional block, we only display the convolutional and max pooling layers. Filters for each convolutional layer represent the number of filters * filter size. Feature maps of each convolutional block represent the number of feature maps * size of each feature map. Conv: convolutional layer; MaxP: max pooling layer; FC: fully connected layer.

### 4.3.2. 3D ROI-based and 3D patch-level CNN

For the 3D ROI-based and 3D patch-level approaches, the chosen CNN architecture, shown in Figure 2, consisted of 4 convolutional blocks (with the same structure as in the 3D subject-level) and 3 FC layers (more architecture details are provided in online supplementary eTable 2).

To extract the 3D patches, a sliding window ($50{\times}50{\times}50$ mm$^3$) without overlap was used to convolve over the entire image, generating 36 patches for each image.

For the 3D ROI-based approach, we chose the hippocampus as a ROI, as done in previous studies. We used a cubic patch ($50{\times}50{\times}50$ mm$^3$) enclosing the left (resp. right) hippocampus. The center of this cubic patch was manually chosen based on the MNI template image (ICBM 2009c nonlinear symmetric template). We ensured visually that this cubic patch included all the hippocampus.

For the 3D patch-level approach, two different training strategies were considered. First, all extracted patches were fitted into a single CNN (denoting this approach as 3D patch-level single-CNN). Secondly, we used one CNN for each patch, resulting in finally 36 (number of patches) CNNs (denoting this approach as 3D patch-level multi-CNN).

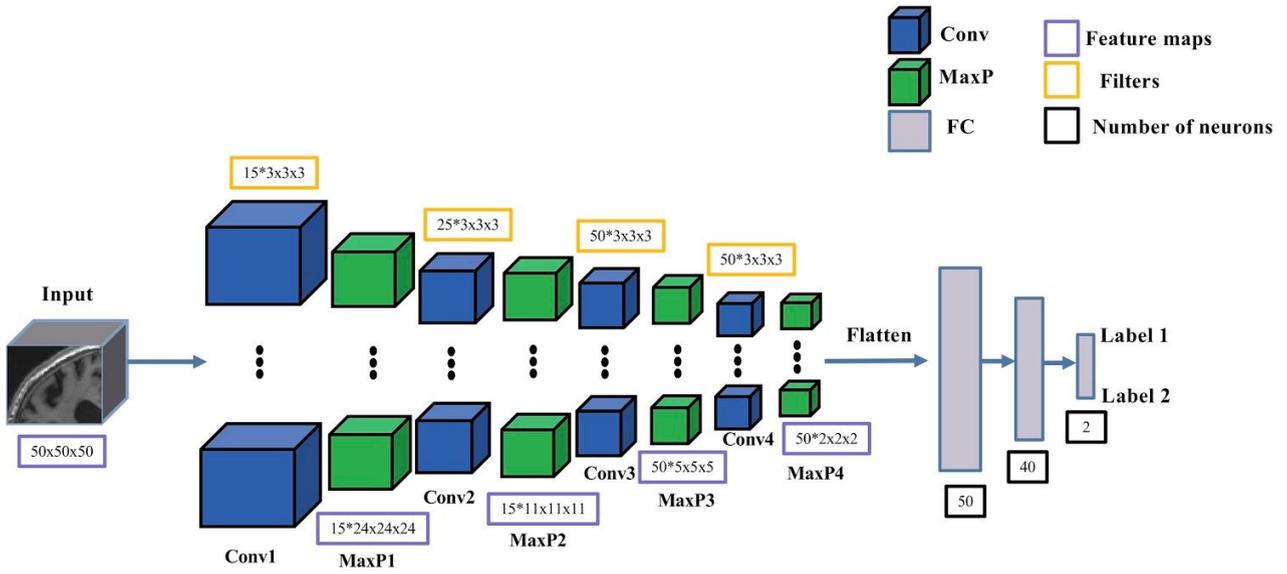

**Figure 2**: Architecture of the 3D ROI-based and 3D patch-level CNNs. For each convolutional block, we only display the convolutional and max pooling layers. Filters for each convolutional layer represent the number of filters * filter size. Feature maps of each convolutional block represent the number of feature maps * size of each feature map. Conv: convolutional layer; MaxP: max pooling layer; FC: fully connected layer.

### 4.3.3. 2D slice-level CNN

For the 2D slice-level approach, the ResNet pre-trained on ImageNet was adopted and fine-tuned. The architecture is shown in Figure 3. The architecture details of ResNet can be found in (He et al., 2016). We added one FC layer on top of the ResNet (more architecture details are provided in online supplementary eTable 3). The last five convolutional layers and the last FC layer of the ResNet, as well as the added FC layer, were fine-tuned. The weight and bias of the other layers of the CNN were frozen during fine-tuning to avoid overfitting.

For each subject, each sagittal slice was extracted and replicated into R, G and B channels respectively, in order to generate a RGB image. The first and last twenty slices were excluded due to the lack of information, which resulted in 129 RGB slices for each image.

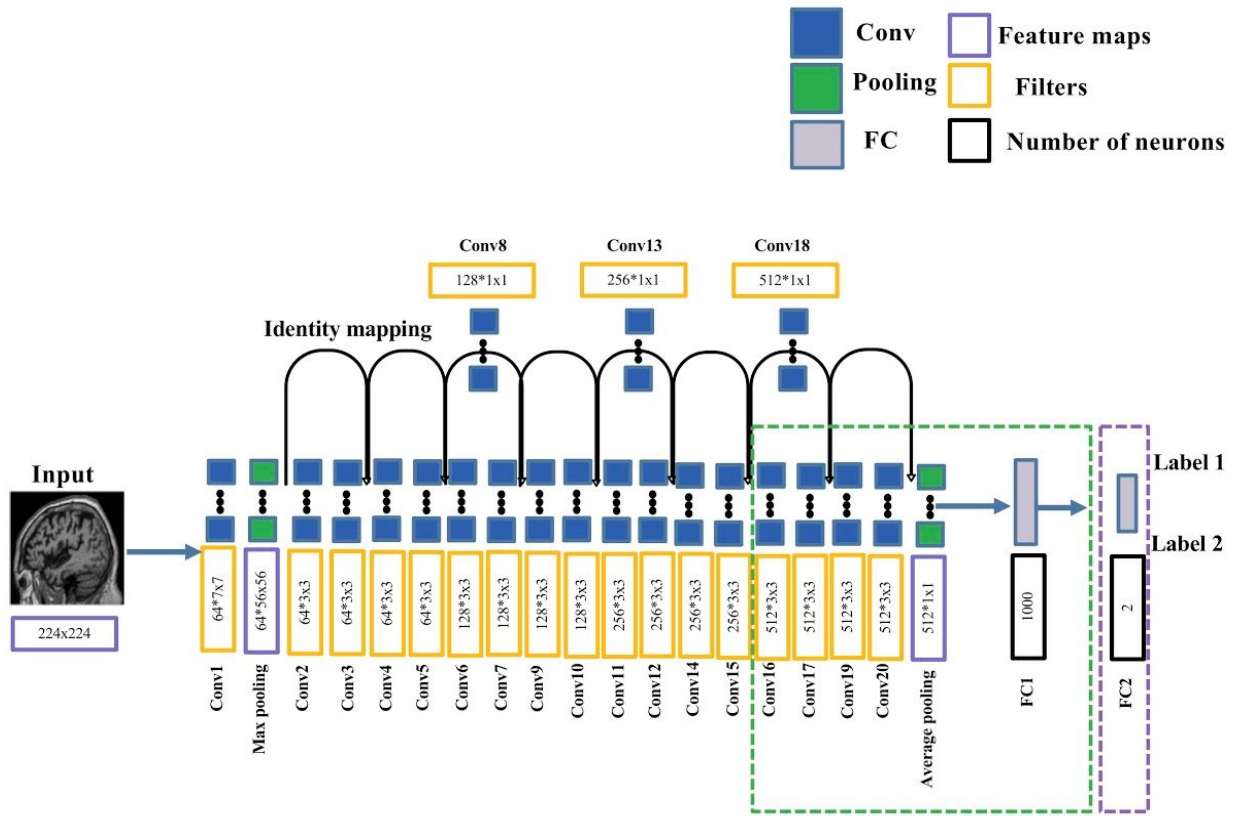

**Figure 3**: Architecture of the 2D slice-level CNN. An FC layer (FC2) was added on top of the ResNet. The last five convolutional layers and the last FC of ResNet (green dotted box) and the added FC layer (purple dotted box) were fine-tuned and the other layers were frozen during training. Filters for each convolutional layer represent the number of filters * filter size. Feature maps of each convolutional block represent the number of feature maps * size of each feature map. Conv: convolutional layer; FC: fully connected layer.

### 4.3.4. Majority voting system

For 3D patch-level, 3D ROI-based and 2D slice-level CNNs, we adopted a soft voting system (Raschka, 2015) to generate the subject-level decision. The subject-level decision is generated based on the decision for each slice (resp. for each patch / for the left and right hippocampus ROI). More precisely, it was computed based on the predicted probability $p$ obtained after softmax normalization of the outputs of all the slices/patches/ROIs/CNNs from the same patient:

$$\hat{y} = arg\ max_i \sum_j^m w_j p_{ij}$$

where $w_j$ is the weight assigned to the j-th patch/slice/ROI/CNN. $w_j$ reflects the importance of each slice/patch/ROI/CNN and is weighted by the normalized accuracy of the j-th slice/patch/ROI/CNN. For the evaluation on the test sets, the weights computed on the validation set were used. Note that the predicted probability $p$ is not calibrated and should be interpreted with care as it is not reflective of the true underlying probability of the sample applied to CNNs (Guo et al., 2017; Kuhn and Johnson, 2013).

For the 3D patch-level multi-CNN approach, the 36 CNNs were trained independently. In this case, the weaker classifiers' weight (balanced accuracy < 0.7) was set to be 0 with the consideration that the labels' probabilities of these classifiers could harm the majority voting system.

### 4.3.5. Comparison to a linear SVM on voxel-based features

For comparison purposes, classification was also performed with a linear SVM classifier. We chose the linear SVM as we previously showed that it obtained higher or at least comparable classification accuracy compared to other conventional models (logistic regression and random forest) (Samper-González et al., 2018). Moreover, given the very high-dimensionality of the input, a nonlinear SVM, e.g. with a radial basis function kernel, may not be advantageous since it would only transport the data into an even higher dimensional space. The SVM took as input the modulated gray matter density maps non-linearly registered to the MNI space using the DARTEL method (Ashburner, 2007), as in our previous study (Samper-González et al., 2018).

## 4.4. Transfer learning

Two different approaches were used for transfer learning: i) AE pre-training for 3D CNNs; and ii) ResNet pre-trained on ImageNet for 2D CNNs.

### 4.4.1. AE pre-training

The AE was constructed based on the architecture of the classification CNN. The encoder part of the AE is composed of a sequence of convolutional blocks, each block having one convolutional layer, one batch normalization layer, one ReLU and one max pooling layer, which is identical to the sequence of convolutional blocks composing the 3D subject-level network. The architecture of the decoder mirrored that of the encoder, except that the order of the convolution layer and the ReLU was swapped. Of note, the pre-training with AE and classification with CNNs in our experiments used the same training and validation data splits in order to avoid potential data leakage problems. Also, each AE was trained on all available data in the training sets. For instance, all MCI, AD and CN subjects in the training dataset were used to pre-train the AE for the AD vs CN classification task.

### 4.4.2. ImageNet pre-training

For the 2D-slice experiments, we investigated the possibility to transfer a ResNet pre-trained on ImageNet (He et al., 2016) to our specific tasks. Next, the fine-tuning procedure was performed on the chosen layers (see Figure 3).

## 4.5. Classification tasks

We performed two tasks in our experiments. AD vs CN was used as the baseline task to compare the results of our different frameworks. Then the best frameworks were selected to perform the prediction task sMCI vs pMCI: the weights and biases of the model learnt on the source task (AD vs CN) were transferred to a new model fine-tuned on the target task (sMCI vs pMCI). For the SVM, the sMCI vs pMCI experiment was performed either by training directly on sMCI vs pMCI or by training on AD vs CN and applying the trained model to sMCI vs pMCI.

## 4.6. Evaluation strategy

### 4.6.1. Validation procedure

Rigorous validation is essential to objectively assess the performance of a classification framework. This is particularly critical in the case of DL as one may easily overfit the validation dataset when manually performing model selection and hyperparameter fine-tuning. An independent test set should be, at the very beginning, generated and concealed. It should not be touched until the CV, based on the training and validation datasets, is finished and the final model is chosen. This test dataset should be used only to assess the performance (i.e. generalization) of a fully specified and trained classifier (Kriegeskorte et al., 2009; Ripley, 1996; Sarle, 1997). Considering this, we chose a classical split into training/validation/test sets. Training/validation sets were used in a CV procedure for model selection while the test set was left untouched until the end of the peer-review process. Only the best performing model for each approach (3D subject-level, 3D patch-level, 3D ROI-based, 2D slice-level), as defined by the CV on training/validation sets, was tested on the test set.

The ADNI test dataset consisted of 100 randomly chosen age- and sex-matched subjects for each diagnostic class (i.e. 100 CN subjects, 100 AD patients). The rest of the ADNI data was used as training/validation set. We ensured that age and sex distributions between training/validation and test sets were not significantly different. Two other test sets were composed of all subjects of OASIS and AIBL. The ADNI test set will be used to assess model generalization within the same dataset (thereby assessing that the model has not overfitted the training/validation set). The AIBL test set will be used to assess generalization to another dataset that has similar inclusion criteria and image acquisition parameters to those of the training set. The OASIS test will be used to assess generalization to a dataset with different inclusion criteria and image acquisition parameters. As mentioned above, it is important to note that the diagnosis labels are not based on the same criteria in OASIS on the one hand and ADNI/AIBL on the other. Thus we do not hypothesize that the models trained on ADNI will generalize well to OASIS.

The model selection procedure, including model architecture selection and training hyperparameter fine-tuning, was performed using only the training/validation dataset. For that purpose, a 5-fold CV was performed, which resulted in one fold (20%) of the data for validation and the rest for training. Note that the 5-fold data split was performed only once for all the experiments with a fixed seed number (*random_state* = 2), thus guaranteeing that all the experiments used exactly the same subjects during CV. Also, no overlapping exists between the MCI subjects used for AE pre-training (using all available AD, CN and MCI) and the test dataset of sMCI vs pMCI. Thus, the evaluation of the cross-task transfer learning (from AD vs CN to sMCI vs pMCI) is unbiased. Finally, for the linear SVM, the hyperparameter C controlling the amount of regularization was chosen using an inner loop of 10-fold CV (thereby performing a nested CV).

The validation procedure includes a series of tests. We implemented tests to check the absence of data leakage in the cross-validation procedure. We also include functional tests of pipelines on inseparable and fully separable data for sanity check. The inseparable data is made as follows. We selected a random subject from OASIS. We then generated multiple subjects by adding random noise to this subject. The images are different but inseparable. Each of the generated subjects was assigned randomly to a diagnostic class. The fully separable data was built as follows. The first (resp. second) group of subjects is made of images in which the voxel intensities of the left (resp. right) hemisphere were lowered. The scripts needed to generate the synthetic datasets are provided in the repository (see https://github.com/aramis-lab/AD-DL).

### 4.6.2. Metrics

We computed the following performance metrics: balanced accuracy (BA), area under the receiver operating characteristic (ROC) curve (AUC), accuracy, sensitivity and specificity. In the manuscript, for the sake of concision, we report only the BA but all other metrics are available on Zenodo under the DOI 10.5281/zenodo.3491003.

## 4.7. Implementation details

The image preprocessing procedures were implemented with Nipype (Gorgolewski et al., 2011). The DL models were built using the Pytorch library[7] (Paszke et al., 2017). TensorboardX[8] was embedded into the current framework to dynamically monitor the training process. Specifically, we evaluated and reported the training and validation BA/loss after each epoch or certain iterations. Of note, instead of using only the current batch of data, the BA was evaluated based on all the training/validation data. Moreover, we organized the classification outputs in a hierarchical way inspired from BIDS, including the TSV files containing the classification results, the outputs of TensorboardX for dynamic monitoring of the training and the best performing models selected based on the validation BA. The linear SVM was implemented using scikit-learn (Pedregosa et al., 2011; Samper-González et al., 2018).

---

[7] https://pytorch.org/
[8] https://github.com/lanpa/tensorboardX

We applied the following early stopping strategy for all the classification experiments: the training procedure does not stop until the validation loss is continuously higher than the lowest validation loss for *N* epochs. Otherwise, the training continues to the end of the pre-defined number of epochs. The selected model was the one which obtained the highest validation BA during training. For the AE pre-training, the AE was trained to the end of the pre-defined number of epochs. We then visually check the validation loss and the quality of the reconstructed images. The mean square loss was used for the AE pre-training and the cross-entropy loss, which combines a log softmax normalization and the negative log likelihood loss, was used for the CNNs.

# 5. Experiments and results

## 5.1. Results on training/validation set

The different classification experiments and results (validation BA during 5-fold CV) are detailed in Table 5. For each experiment, the training process of the best fold (with highest balanced validation accuracy) is presented as an illustration (see supplementary eFigures 1-4 for details). Lastly, the training hyperparameters (e.g. learning rate and batch size) for each experiment are presented in supplementary eTable 4.

All the pipelines (3D subject-level, 3D ROI-based, 3D patch-level, 2D slice-level and SVM) were tested on the synthetic inseparable and fully separable datasets. The results were as expected: 0.5 (resp. 1.00) of balanced accuracy for the inseparable (resp. fully separable) dataset.

### 5.1.1. 3D subject-level

**Influence of intensity rescaling.** We first assessed the influence of intensity rescaling. Without rescaling, the CNN did not perform better than chance (BA = 0.50) and there was an obvious generalization gap (high training but low validation BA). With intensity rescaling, the BA improved to 0.80. Based on these results, intensity rescaling was used in all subsequent experiments.

**Influence of transfer learning (AE pre-training).** The performance was slightly higher with AE pre-training (0.82) than without (0.80). Based on this, we decided to always use AE pre-training, even though the difference is small.

**Influence of the training dataset size.** We then assessed the influence of the amount of training data, comparing training using only baseline data to those with longitudinal data. The performance was moderately higher with longitudinal data (0.85) compared to baseline data only (0.82). We choose to continue exploring the influence of this choice because the four different approaches have a very different number of learnt parameters and the sample size is intrinsically augmented in 2D slice-level and 3D single-CNN patch-level approaches.

**Influence of preprocessing.** We then assessed the influence of the preprocessing comparing the "Extensive" and "Minimal" preprocessing procedures. The performance was almost equivalent with the "Minimal" preprocessing (0.85) and with the "Extensive" preprocessing (0.86). Hence in the following experiments we kept the "Minimal" preprocessing.

**Classification of sMCI vs pMCI.** The BA was the same for baseline data and for longitudinal data (0.73).

### 5.1.2.    3D ROI-based

For AD vs CN, the BA was 0.88 for baseline data and 0.86 for longitudinal data. This is slightly higher than that of the subject-level approach. For sMCI vs pMCI, the BA was 0.77 for baseline data and 0.78 for longitudinal data. This is substantially higher than with the 3D-subject level approach.

### 5.1.3.    3D patch-level

**Single CNN.** For AD vs CN, the BA was 0.74 for baseline data and 0.76 for longitudinal data.

**Multi CNN.** For AD vs CN, the BA was 0.81 for baseline data and 0.83 for longitudinal data, thereby outperforming the single CNN approach. For sMCI vs pMCI, the BA was 0.75 for baseline data and 0.77 for longitudinal data. The performance for both tasks is slightly lower than that of the 3D ROI-based approach. Compared to the 3D subject-level approach, this method works better for sMCI vs pMCI.

### 5.1.4.    2D slice-level

In general, the performance of the 2D-slice level approach was lower to that of the 3D ROI-based, 3D patch-level multi CNN and 3D subject-level (when trained with longitudinal data) approaches but higher than that of the 3D patch-level single CNN approach. For 2D slice-level, the use of longitudinal data for training did not improve the performance (0.79 for baseline data; 0.74 for longitudinal data). Finally, we studied the influence of data leakage using a slice-level data split strategy. As expected, the BA was 1.00.

### 5.1.5.    Linear SVM

For task AD vs CN, the balanced accuracies were 0.88 when trained with baseline data and 0.87 when trained with longitudinal data. For task sMCI vs pMCI, when training from scratch, the balanced accuracies were 0.68 when trained with baseline data and 0.68 when trained with longitudinal data. When using transfer learning from the task AD vs CN to the task sMCI vs pMCI, the balanced accuracies were 0.70 (when trained with baseline data) and 0.70 (when trained with longitudinal data). The performance of the SVM on AD vs CN is thus higher than that of most DL models and comparable to the best ones. Whereas for task sMCI vs pMCI, the BA of the SVM is lower than that of DL models.

**Table 5.** Summary of all the classification experiments and validation results in our analyses. For each model, we report the balanced accuracy for each of the five folds within brackets and the average and standard-deviation across the folds. Note that this is not the standard-deviation of the estimator of balanced accuracy.

MinMax: for CNNs, intensity rescaling was done based on min and max values, resulting all values to be in the range of [0, 1]; SPM-based: the SPM-based gray matter maps are intrinsically rescaled; AE: autoencoder. For DL models, sMCI vs pMCI tasks were done with as follows: the weights and biases of the model learnt on the source task (AD vs CN) were transferred to a new model fine-tuned on the target task (sMCI vs pMCI). For SVM, the sMCI vs pMCI was done either training directly on sMCI vs pMCI or using training on AD vs CN and applying the trained model to sMCI vs pMCI.

| Classification architectures | Training data | Image preprocessing | Intensity rescaling | Data split | Training approach | Transfer learning | Task | Validation balanced accuracy | Exp # |
|---|---|---|---|---|---|---|---|---|---|
| 3D subject-level CNN | Baseline | Minimal | None | subject-level | single-CNN | None | AD vs CN | $0.50 \pm 0.00$ [0.50, 0.50, 0.50, 0.50, 0.50] | 1 |
| | | | MinMax | | | | | $0.80 \pm 0.05$ [0.76, 0.86, 0.81, 0.85, 0.74] | 2 |
| | | | | | | AE pre-train | | $0.82 \pm 0.05$ [0.74, 0.90, 0.83, 0.77, 0.83] | 3 |
| | Longitudinal | Minimal | MinMax | subject-level | single-CNN | AE pre-train | | $0.85 \pm 0.04$ [0.88, 0.88, 0.84, 0.85, 0.78] | 4 |
| | | Extensive | | | | | | $0.86 \pm 0.06$ [0.88, 0.94, 0.85, 0.85, 0.76] | 5 |
| | | Minimal | | | | | sMCI vs pMCI | $0.73 \pm 0.03$ [0.73, 0.73, 0.67, 0.76, 0.74] | 6 |
| | Baseline | | | | | | | $0.73 \pm 0.05$ [0.73, 0.73, 0.63, 0.77, 0.76] | 7 |

| Classification architectures | Training data | Image preprocessing | Intensity rescaling | Data split | Training approach | Transfer learning | Task | Validation balanced accuracy | Exp # |
|---|---|---|---|---|---|---|---|---|---|
| 3D ROI-based CNN | Baseline | Minimal | MinMax | subject-level | single-CNN | AE pre-train | AD vs CN | 0.88 ± 0.03 [0.84, 0.89, 0.90, 0.89, 0.85] | 8 |
| | | | | | | | sMCI vs pMCI | 0.77 ± 0.05 [0.81, 0.81, 0.67, 0.78, 0.76] | 9 |
| | Longitudinal | | | | | | AD vs CN | 0.86 ± 0.02 [0.83, 0.86, 0.86, 0.88, 0.86] | 10 |
| | | | | | | | sMCI vs pMCI | 0.78 ± 0.07 [0.87, 0.73, 0.68, 0.82, 0.78] | 11 |
| 3D patch-level CNN | Baseline | Minimal | MinMax | subject-level | single-CNN | AE pre-train | AD vs CN | 0.74 ± 0.08 [0.75, 0.84, 0.78, 0.75, 0.59] | 12 |
| | Longitudinal | | | | | | | 0.76 ± 0.04 [0.78, 0.77, 0.80, 0.78, 0.69] | 13 |
| | Baseline | | | | multi-CNN | | AD vs CN | 0.81 ± 0.03 [0.82, 0.84, 0.83, 0.77, 0.79] | 14 |
| | | | | | | | sMCI vs pMCI | 0.75 ± 0.04 [0.80, 0.72, 0.72, 0.79, 0.72] | 15 |
| | Longitudinal | | | | | | AD vs CN | 0.83 ± 0.02 [0.83, 0.85, 0.84, 0.82, 0.79] | 16 |
| | | | | | | | sMCI vs pMCI | 0.77 ± 0.04 [0.77, 0.75, 0.71, 0.82, 0.79] | 17 |

| Classification architectures | Training data | Image preprocessing | Intensity rescaling | Data split | Training approach | Transfer learning | Task | Validation balanced accuracy | Exp # |
|---|---|---|---|---|---|---|---|---|---|
| 2D slice-level CNN | Baseline | Minimal | MinMax | subject-level | single-CNN | ImageNet pre-train | AD vs CN | 0.79 ± 0.04 [0.83, 0.83, 0.72, 0.82, 0.73] | 18 |
| | Longitudinal | | | | | | | 0.74 ± 0.03 [0.76, 0.80, 0.74, 0.71, 0.69] | 19 |
| | Baseline | | | slice-level (data leakage) | | | | 1.00 ± 0 [1.00, 1.00, 1.00, 1.00, 1.00] | 20 |
| SVM | Baseline | DartelGM | SPM-based | subject-level | None | None | AD vs CN | 0.88 ± 0.02 [0.92, 0.89, 0.85, 0.89, 0.84] | 21 |
| | | | | | | | sMCI vs pMCI (trained on sMCI vs pMCI) | 0.68 ± 0.02 [0.71, 0.68, 0.66, 0.67, 0.71] | 22 |
| | | | | | | | sMCI vs pMCI (trained on AD vs CN) | 0.70 ± 0.06 [0.66, 0.75, 0.70, 0.79, 0.63] | 23 |
| | | | | | | None | AD vs CN | 0.87 ± 0.01 [0.86, 0.86, 0.88, 0.87, 0.85] | 24 |
| | Longitudinal | | | | | | sMCI vs pMCI (trained on sMCI vs pMCI) | 0.68 ± 0.06 [0.75, 0.77, 0.62, 0.62, 0.67] | 25 |
| | | | | | | | sMCI vs pMCI (trained on AD vs CN) | 0.70 ± 0.02 [0.68, 0.72, 0.67, 0.69, 0.73] | 26 |

## 5.2. Results on the test sets

Results on the three test sets (ADNI, OASIS and AIBL) are presented in Table 6. For each category of approach, we only applied the best models for both baseline and longitudinal data.

### 5.2.1. 3D subject-level

For AD vs CN, all models generalized well to the ADNI and AIBL test sets but not to the OASIS test set (losing over 0.15 points of BA).

For sMCI vs pMCI, the models generalized relatively well to the ADNI test set but not to the AIBL test set (losing over 0.20 points). Note that the generalization was better for longitudinal than for baseline.

### 5.2.2. 3D ROI-based

For AD vs CN, the models generalized well to the ADNI test set, slightly worse to the AIBL test set (losing 0.04 to 0.05 points) and considerably worse for OASIS (losing from 0.13 to 0.19 points).

For sMCI vs pMCI, there was a slight decrease in BA on the ADNI test set and a severe decrease for the AIBL test set. Note that on the ADNI test set, the performance of the 3D ROI-based is almost the same as that of the 3D-subject (when using longitudinal data) while it was better on the validation set.

### 5.2.3. 3D patch-level

For AD vs CN, the generalization pattern was similar to that of the other models: good for ADNI and AIBL, poor for OASIS.

For sMCI vs pMCI, the BA on the ADNI test set was 0.05 to 0.07 points lower than on the ADNI validation set. The BA on the AIBL test set was very poor.

### 5.2.4. 2D slice-level

For AD vs CN, there was a slight decrease in performance on the ADNI test set (losing from 0 to 0.03 points) and the AIBL test set (losing from 0.01 to 0.03 points) and a considerable decrease on the OASIS test set (losing from 0.13 to 0.14 points). As expected, the "data-leakage" model did not generalize well.

### 5.2.5. Linear SVM

For AD vs CN, we observed the same pattern as for the other models: excellent generalization to ADNI and AIBL but not to OASIS.

For sMCI vs pMCI, the generalization was excellent for ADNI but not for AIBL. Of note, the BA on the ADNI test set was even higher to that of the validation, reaching a level which is comparable to the best DL models.

**Table 6.** Summary of the results of the three test datasets in our analyses. 3D subject-level CNNs were trained using intensity rescaling and our "Minimal" preprocessing, with a data split on the subject level and transfer learning (AE pretraining for AD vs CN tasks and cross-task transfer learning was applied for sMCI vs pMCI tasks). For each model, we first copied the validation balanced accuracy (averaged across the five folds) that is reported in Table 5. Then, we report the balanced accuracy for each test set (ADNI, AIBL, OASIS), more specifically within brackets we report the balanced accuracy for each of the trained models of the 5 folds of the validation set and then the average across the five folds.

MinMax: for CNNs, intensity rescaling was done based on min and max values, resulting all values to be in the range of [0, 1]; SPM-based: the SPM-based gray matter maps are intrinsically rescaled; AE: autoencoder.

| Classification architectures | Training data | Image preprocessing | Intensity rescaling | Data split | Training approach | Transfer learning | Task | Validation balanced accuracy | ADNI test balanced accuracy | AIBL test balanced accuracy | OASIS test balanced accuracy |
|---|---|---|---|---|---|---|---|---|---|---|---|
| 3D subject-level CNN | Baseline | Image processing = Minimal | MinMax | subject-level | single-CNN | AE pre-train | AD vs CN | $0.82 \pm 0.05$ | 0.82 [0.79, 0.85, 0.82, 0.81, 0.85] | 0.83 [0.81, 0.85, 0.84, 0.78, 0.86] | 0.67 [0.59, 0.69, 0.72, 0.64, 0.69] |
| | Longitudinal | | | | | | | $0.85 \pm 0.04$ | 0.85 [0.88, 0.84, 0.84, 0.84, 0.84] | 0.86 [0.89, 0.85, 0.86, 0.85, 0.86] | 0.68 [0.65, 0.70, 0.70, 0.71, 0.65] |
| | Baseline | | | | | | sMCI vs pMCI | $0.73 \pm 0.05$ | 0.69 [0.68, 0.71, 0.64, 0.73, 0.67] | 0.52 [0.51, 0.47, 0.55, 0.54, 0.55] | -- |
| | Longitudinal | | | | | | | $0.73 \pm 0.03$ | 0.73 [0.75, 0.72, 0.72, 0.74, 0.72] | 0.50 [0.48, 0.47, 0.54, 0.52, 0.51] | -- |

| Classification architectures | Training data | Image preprocessing | Intensity rescaling | Data split | Training approach | Transfer learning | Task | Validation balanced accuracy | ADNI test balanced accuracy | AIBL test balanced accuracy | OASIS test balanced accuracy |
|---|---|---|---|---|---|---|---|---|---|---|---|
| 3D ROI-based CNN | Baseline | Minimal | MinMax | subject-level | single-CNN | AE pre-train | AD vs CN | 0.88 ± 0.03 | 0.89 [0.87, 0.88, 0.90, 0.91, 0.89] | 0.84 [0.83, 0.88, 0.84, 0.85, 0.83] | 0.69 [0.62, 0.74, 0.70, 0.69, 0.71] |
| | | | | | | | sMCI vs pMCI | 0.77 ± 0.05 | 0.74 [0.75, 0.72, 0.76, 0.75, 0.75] | 0.60 [0.56, 0.56, 0.66, 0.62, 0.59] | -- |
| | Longitudinal | | | | | | AD vs CN | 0.86 ± 0.02 | 0.85 [0.87, 0.82, 0.87, 0.86, 0.87] | 0.81 [0.79, 0.81, 0.79, 0.82, 0.85] | 0.73 [0.71, 0.73, 0.72, 0.76, 0.71] |
| | | | | | | | sMCI vs pMCI | 0.78 ± 0.07 | 0.74 [0.70, 0.73, 0.73, 0.75, 0.81] | 0.57 [0.56, 0.53, 0.52, 0.66, 0.56] | -- |
| 3D patch-level CNN | Baseline | Minimal | MinMax | subject-level | multi-CNN | AE pre-train | AD vs CN | 0.81 ± 0.03 | 0.81 [0.82, 0.81, 0.84, 0.80, 0.79] | 0.81 [0.81, 0.75, 0.81, 0.84, 0.82] | 0.64 [0.61, 0.65, 0.60, 0.69, 0.67] |
| | | | | | | | sMCI vs pMCI | 0.75 ± 0.04 | 0.70 [0.71, 0.66, 0.66, 0.71, 0.75] | 0.64 [0.63, 0.52, 0.67, 0.74, 0.63] | -- |
| | Longitudinal | | | | | | AD vs CN | 0.83 ± 0.02 | 0.86 [0.86, 0.86, 0.87, 0.85, 0.84] | 0.80 [0.82, 0.78, 0.81, 0.81, 0.79] | 0.71 [0.70, 0.70, 0.71, 0.71, 0.67] |
| | | | | | | | sMCI vs pMCI | 0.77 ± 0.04 | 0.70 [0.70, 0.71, 0.69, 0.71, 0.69] | 0.44 [0.45, 0.39, 0.55, 0.42, 0.39] | -- |

| Classification architectures | Training data | Image preprocessing | Intensity rescaling | Data split | Training approach | Transfer learning | Task | Validation balanced accuracy | ADNI test balanced accuracy | AIBL test balanced accuracy | OASIS test balanced accuracy |
|---|---|---|---|---|---|---|---|---|---|---|---|
| 2D slice-level CNN | Baseline | Minimal | MinMax | subject-level | single-CNN | ImageNet pre-train | AD vs CN | 0.79 ± 0.04 | 0.76 [0.76, 0.75, 0.77, 0.75, 0.78] | 0.76 [0.74, 0.76, 0.78, 0.75, 0.75] | 0.65 [0.67, 0.62, 0.64, 0.65, 0.69] |
|  | Longitudinal |  |  |  |  |  |  | 0.74 ± 0.03 | 0.74 [0.81, 0.76, 0.70, 0.74, 0.72] | 0.73 [0.72, 0.77, 0.72, 0.66, 0.79] | 0.61 [0.62, 0.63, 0.64, 0.58, 0.60] |
|  | Baseline |  |  | slice-level (data leakage) |  |  |  | 1.00 ± 0 | 0.75 [0.74, 0.76, 0.75, 0.76, 0.75] | 0.80 [0.80, 0.79, 0.82, 0.80, 0.81] | 0.68 [0.68, 0.67, 0.69, 0.70, 0.66] |
| SVM | Baseline | DartelGM | SPM-based | subject-level | None | None | AD vs CN | 0.88 ± 0.02 | 0.88 [0.88, 0.87, 0.90, 0.90, 0.88] | 0.88 [0.87, 0.90, 0.87, 0.89, 0.90] | 0.70 [0.71, 0.71, 0.70, 0.68, 0.72] |
|  |  |  |  |  |  |  | sMCI vs pMCI (trained on AD vs CN) | 0.70 ± 0.06 | 0.75 [0.75, 0.75, 0.74, 0.76, 0.76] | 0.60 [0.62, 0.54, 0.62, 0.59, 0.64] | -- |
|  | Longitudinal |  |  |  |  |  | AD vs CN | 0.87 ± 0.01 | 0.87 [0.85, 0.84, 0.90, 0.89, 0.87] | 0.87 [0.88, 0.86, 0.88, 0.87, 0.89] | 0.71 [0.73, 0.68, 0.72, 0.70, 0.71] |
|  |  |  |  |  |  |  | sMCI vs pMCI (trained on AD vs CN) | 0.70 ± 0.02 | 0.76 [0.74, 0.75, 0.80, 0.77, 0.76] | 0.68 [0.67, 0.66, 0.68, 0.67, 0.71] | -- |

## 6. Discussion

The present study contains three main contributions. First, we performed a systematic and critical literature review, which highlighted several important problems. Then, we proposed an open-source framework for the reproducible evaluation of AD classification using CNNs and T1w MRI. Finally, we applied the framework to rigorously compare different CNN approaches and to study the impact of key components on the performance. We hope that the present paper will provide a more objective assessment of the performance of CNNs for AD classification and constitute a solid baseline for future research.

This paper first proposes a survey of existing CNN methods for AD classification that highlighted several serious problems with the existing literature. We found that data leakage was potentially present in half of the 32 surveyed studies. This problem was evident in six of them and possible (due to inadequate description of the validation procedure) in ten others. This is a very serious issue, in particular considering that all these studies have undergone peer-review, likely to bias the performance upwards. We confirmed this assumption by simulating data leakage and found that it led to a biased evaluation of the BA (1.00 on the validation instead of 0.75 on ADNI test set and 0.80 on AIBL test set). Similar findings were observed in (Bäckström et al., 2018). Moreover, the survey highlighted that many studies did not motivate the choice of their architecture or training hyperparameters. Only two of them (Wang et al., 2019; S.-H. Wang et al., 2018) explored and gave results obtained with different architecture hyperparameters. However, it is possible that these results were computed on the test set to help choose their final model, hence they may be contaminated by data leakage. For other studies, it is also likely that multiple combinations of architecture and training hyperparameters were tested, leading to a biased performance on the test set. We believe that these issues may potentially be caused by the lack of expertise in medical imaging or DL. For instance, splitting at the slice-level comes from a lack of knowledge of the nature of medical imaging data. We hope that the present paper will help to spread knowledge and good practices in the field.

The second contribution of our work is an open-source framework for reproducible experiments on AD classification using CNNs. Some studies in our bibliography made their code available on open source platforms (Hon and Khan, 2017; Hosseini-Asl et al., 2016; Korolev et al., 2017; Manhua Liu et al., 2018). Even though this practice should be encouraged, it does not guarantee reproducibility of the results. Two studies (Cheng and Liu, 2017; Manhua Liu et al., 2018) used the online code of (Hosseini-Asl et al., 2016) for comparison with their framework, but neither of them succeeded in reproducing the results of the original study (for the AD vs CN task they report both an accuracy of 0.82 while the original study reports an accuracy of 0.99). We extended our open-source framework for reproducible evaluation of AD classification, initially dedicated to traditional methods (Samper-González et al., 2018; J. Wen et al., 2018), to DL approaches. It is composed of the previously developed tools for data management that rely on the BIDS community standard (Gorgolewski et al., 2016), a new image preprocessing pipeline performing bias field correction, affine registration to MNI space and intensity rescaling, a set of CNN models that are representative of the literature, and rigorous validation procedures dedicated to DL. We hope that this open-source framework will facilitate the reproducibility and objectivity of DL methods for AD classification as it enables researchers to easily embed new image preprocessing pipelines or CNN architectures and study their added value. It extends the efforts initiated in both the neuroimaging (Gorgolewski and Poldrack, 2016; Poldrack et al., 2017) and ML (Sonnenburg et al., 2007; Stodden et al., 2014; Vanschoren et al., 2014) communities to improve reproducibility. In particular, frameworks and software tools have been recently proposed to facilitate and standardize ML analyses for neuroimaging data. Nilearn[9] is currently mostly focused on fMRI data. It provides pipelines for image processing, various techniques for decoding activity and studying functional connectivity as well as visualization tools. Neuropredict[10] (Raamana, 2017) is more focused on computer-aided diagnosis and other clinical applications. In particular, it provides standardized cross-validation procedures and tools to visualize results. .

Our third contribution is the rigorous assessment of the performance of different CNN architectures. The proposed framework was applied to images from three public datasets, ADNI, AIBL and OASIS. On the ADNI test dataset, the diagnostic BA of CNNs ranged from 0.76 to 0.89 for the AD vs CN task and from 0.69 to 0.74 for the sMCI vs pMCI task. These results are in line with the state-of-the-art (studies without data leakage in Table 1A), where classification accuracy typically ranged from 0.76 to 0.91 for AD vs CN and 0.62 to 0.83 for sMCI vs pMCI. Nevertheless, the performance that we report is lower than that of the top-performing studies. This potentially comes from the fact that our test set was fully independent and was never used to choose the architectures or parameters. The proposed framework can be used to provide a baseline performance when developing new methods.

---

[9] https://nilearn.github.io/
[10] http://github.com/raamana/neuropredict

Different approaches, namely 3D subject-level, 3D ROI-based, 3D patch-level and 2D slice-level CNNs, were compared. Our study is the first one to systematically compare the performance of these four approaches. In the literature, three studies (Cheng et al., 2017; Li et al., 2018; Manhua Liu et al., 2018) using a 3D patch-level approach compared their results with a 3D subject-level approach. In all studies, the 3D patch-level multi-CNN gave better results than the 3D-subject CNN (3 or 4 percent points of difference between the two approaches). However, except for (Manhua Liu et al., 2018) where the code provided by (Hosseini-Asl et al., 2016) is reused, the methods used for the comparison are poorly described and the studies would thus be difficult, if not impossible, to reproduce. In general, in our results, three approaches (3D subject-level, 3D ROI-based, 3D patch-level) provided approximately the same level of performance (note that this discussion paragraph is based on test set results which are the most objective performance measures). On the other hand, the 2D-slice approach was less efficient. One can hypothesize that this is because the spatial information is not adequately modeled by these approaches (no 3D consistency across slices). Only one paper (without data leakage) has explored 2D slice-level using ImageNet pre-trained ResNet (Valliani and Soni, 2017). Their accuracy is very similar to ours (0.81 for task AD vs CN). The results of the three 3D approaches were in general comparable and our results do not allow a strong conclusion to be drawn on the superiority of one of these three approaches. Nevertheless, there is a trend for a slightly lower performance of the 3D patch-level approach. It could come from the fact that the spatial information is also not ideally modeled with the 3D patch (no consistency at the border of the patch). Other studies with 3D patch-level approaches in the literature (Cheng et al., 2017; Lian et al., 2018; Li et al., 2018; Mingxia Liu et al., 2018a, 2018c) reported higher accuracies (from 0.87 to 0.91) than ours (from 0.81 to 0.86). We hypothesize that it may come from the increased complexity of their approach, including patch selection and fusion. Concerning the 3D ROI-based approach, two papers in the literature using hippocampal ROI reported high accuracies for task AD vs CN (0.84 and 0.90), comparable to ours, even though their definition of the ROI was different (Aderghal et al., 2018, 2017b). As for the 3D subjects (Bäckström et al., 2018; Cheng and Liu, 2017; Korolev et al., 2017; Li et al., 2017; Senanayake et al., 2018; Shmulev et al., 2018), results of the literature varied across papers, from 0.76 to 0.90. Although we cannot prove it directly, we believe that this variability stems from the high risk of overfitting. To summarize, our results demonstrate the superiority of 3D approaches compared to 2D, but the results of the different 3D approaches were not substantially different. In light of this, one could prefer using the 3D ROI-based method which requires less memory and training time (compared to other 3D methods) and is conceptually simpler than the 3D-patch multi-CNN approach. However, it could be that future works, with larger training sets, result in the superiority of approaches that exploit all the information in the 3D image and not only that of the hippocampus. Indeed, even though the hippocampus is affected early and severely by AD (Braak and Braak, 1998), alterations in AD are not confined to the hippocampus and extend to other regions in the temporal, parietal and frontal lobes.

One interesting question is whether DL could perform better than conventional ML methods for AD classification. Here, we chose to compare CNN to a linear SVM. SVM has been used in many AD classification studies and obtained competitive balanced accuracies (Falahati et al., 2014; Haller et al., 2011; Rathore et al., 2017). In the current study, the SVM was at least as good as the best CNNs for both the AD vs CN and the sMCI vs pMCI task. Note that we used a standard linear SVM with standard voxel-based features. It could be that more sophisticated conventional ML methods could provide even higher performance. Similarly, we do not claim that more sophisticated DL architectures would not outperform the SVM. However, this is not the case with the architectures that we tested, which are representative of the existing literature on AD classification. Besides, it is possible that CNNs will outperform SVM when larger public datasets will become available. Overall, a major result of the present paper is that, with the sample size which is available in ADNI, CNNs did not provide an increase in performance compared to SVM.

Unbiased evaluation of the performance is an essential task in ML. This is particularly critical for DL because of the extreme flexibility of the models and of the numerous architecture and training hyperparameters that can be chosen. In particular, it is crucial that such choices are not made using the test set. We chose a very strict validation strategy in that respect: the test sets were left untouched until the end of the peer-review process. This guarantees that only the final models, after all possible adjustments, are carried to the test set. Moreover, it is important to assess generalization not only to unseen subjects but also to other studies in which image acquisitions or patient inclusion criteria can vary. In the present paper, we used three test sets from the ADNI, AIBL and OASIS databases to assess different generalization aspects.

We studied generalization in three different settings: i) on a separate test set from ADNI, thus from the same study as those of the training set; ii) on AIBL, i.e. a different study but with similar inclusion criteria and imaging acquisitions; iii) on OASIS, i.e. a study with different inclusion criteria and imaging acquisitions. Overall, the models generalized well to ADNI (for both tasks) and to AIBL (for AD vs CN). On the other hand, we obtained a very poor generalization to sMCI vs pMCI for AIBL. We hypothesize that it could be because pMCI and sMCI participants from AIBL are substantially older than those of ADNI, which is not the case for AD and CN participants. Nevertheless, note that the sample size for sMCI vs pMCI in AIBL is quite small (33 participants). Also, the generalization to OASIS was poor. This may stem from the diagnosis criteria which are less rigorous (in OASIS, all participants with CDR>0 are considered AD). Overall, these results bring important information. First, good generalization to unseen, similar, subjects demonstrate that the models did not overfit the subjects at hand in the training/validation set. On the other hand, poor generalization to different age ranges, protocols and inclusion criteria show that trained models are too specific of these characteristics. Generalization across different populations thus remains an unsolved problem and will require training on more representative datasets but maybe also new strategies to make training more robust to heterogeneity. This is critical for the future translation to clinical practice in which conditions are much less controlled than in research datasets like ADNI.

We studied the influence of several key choices on the performance. First, we studied the influence of AE pre-training and showed that it slightly improved the average over training from scratch. Three previous papers studied the impact of AE pre-training (Hosseini-Asl et al., 2016; Vu et al., 2018, 2017) and found that it improved the results. However, they are all suspected of data leakage. We thus conclude that, to date, it is not proven that AE pre-training leads to a significant increase in BA. A difficulty in AD classification using DL is the limited amount of data samples available for training. However, training with longitudinal instead of baseline data gave only a slight increase of BA in most approaches. The absence of a major improvement may be due to several factors. First, training with longitudinal data implies training with data from more advanced disease stages, since patients are seen at a later point in the disease course. This may have an adverse effect on the performance of the model when tested on baseline data, at which the patients are less advanced. Also, since the additional data come from the same patients, this does not provide a better coverage of inter-individual variability. We studied the impact of image preprocessing. First, as expected, we found that CNNs cannot be successfully trained without intensity rescaling. We then studied the influence of two different preprocessing procedures ("Minimal" and "Extensive"). The "Minimal" procedure is limited to an affine registration of the subject's image to a standard space, while for the "Extensive" procedure non-linear registration and skull stripping are performed. They led to comparable results. In principle, this is not surprising as DL methods do not require extensive preprocessing. In the litterature, varied types of preprocessing have been used. Some studies used non-linear registration (Bäckström et al., 2018; Basaia et al., 2019; Lian et al., 2018; Lin et al., 2018; Mingxia Liu et al., 2018a, 2018c; Wang et al., 2019; S.-H. Wang et al., 2018) while others used only linear (Aderghal et al., 2018, 2017a, 2017b; Hosseini Asl et al., 2018; Li et al., 2018; Manhua Liu et al., 2018; Shmulev et al., 2018) or no registration (Cheng and Liu, 2017). None of them compared these different preprocessings with the exception of (Bäckström et al., 2018) which compared preprocessing using FreeSurfer to no preprocessing. They found that training the network with the raw data resulted in a lower classification performance (drop in accuracy of 38 percent points) compared to the preprocessed data using FreeSurfer (Bäckström et al., 2018). However, FreeSurfer comprises a complex pipeline with many preprocessing steps so it is unclear, from their results, which part drives the superior performance. We clearly demonstrated that the intensity rescaling is essential for the CNN training whereas there is no improvement in using a non-linear registration over a linear one. Finally, we found that, for the 3D-patch level framework, the multi-CNN approach gave better results than the single-CNN one. However, this may be mainly because the multi-CNN approach benefits from a thresholding system which excludes the worst patches, a system that was not present in the single-CNN approach. To test this hypothesis, we performed supplementary experiments in which the multi-CNN was trained without threshold and the single-CNN was trained using the same thresholding system as in the main experiments of the multi-CNN. Results are reported in eTables 6 and 7. We observed that the results of the multi-CNN and the single-CNN are comparable when they use the same thresholding system. For example, for the AD vs CN task, without thresholding, the BA of the multi-CNN was 0.76 using baseline data and 0.72 using longitudinal data while that of the single-CNN were respectively 0.74 and 0.76. A similar observation can be made when both approaches used the thresholding. These supplementary experiments suggest that, under similar conditions, the multi-CNN architecture does not always perform better than the single-CNN architecture. In light of this, it would seem preferable to choose a framework that offers a better compromise between performance and conceptual complexity, *e.g.* the 3D-ROI or the 3D-subject approaches.

Our study has the following limitations. First, a large number of options exist when choosing the model architecture and training hyperparameters. Even though we did our best to make meaningful choices and test a relatively large number of possibilities, we cannot exclude that other choices could have led to better results. To overcome this limitation, our framework is freely available to the community. Researchers can use it to propose and validate potentially better performing models. In particular, with our proposed framework, researchers can easily try their own models without touching the test datasets. Secondly, the CV procedures were performed only once. Of course, the training is not deterministic and one would ideally want to repeat the CV to get a more robust estimate of the performance. However, we did not perform this due to limited computational resources. Finally, overfitting always exists in our experiments, even though different techniques have been tried (e.g. transfer learning, dropout or weight decay). This phenomenon occurs mainly due to the limited size of the datasets available for AD classification. It is likely that training with much larger datasets would result in higher performance.

# Acknowledgements

This work was granted access to the HPC resources of IDRIS under the allocation 2019-100963 made by GENCI (Grand Équipement National de Calcul Intensif) in the context of the Jean Zay "Grands Challenges" (2019). We thank Mr. Maxime Kermarquer for the IT support during this study. We also thank the following colleagues for useful discussions and suggestions: Alexandre Bône and Johann Faouzi. The research leading to these results has received funding from the program "Investissements d'avenir" ANR-10-IAIHU-06 (Agence Nationale de la Recherche-10-Investissements Avenir Institut Hospitalo-Universitaire-6), from ANR-19-P3IA-0001 (Agence Nationale de la Recherche - 19 - ProgrammeInstitutsInterdisciplinairesIntelligenceArtificielle-0001, project PRAIRIE), from the European Union H2020 program (project EuroPOND, grant number 666992), and from the joint NSF/NIH/ANR program "Collaborative Research in Computational Neuroscience" (project HIPLAY7, grant number ANR-16-NEUC-0001-01). J.W. receives financial support from the China Scholarship Council (CSC). O.C. is supported by a "Contrat d'Interface Local" from Assistance Publique-Hôpitaux de Paris (AP-HP).

Data collection and sharing for this project was funded by the Alzheimer's Disease Neuroimaging Initiative (ADNI) (National Institutes of Health Grant U01 AG024904) and DOD ADNI (Department of Defense award number W81XWH-12-2-0012). ADNI is funded by the National Institute on Aging, the National Institute of Biomedical Imaging and Bioengineering, and through generous contributions from the following: AbbVie, Alzheimer's Association; Alzheimer's Drug Discovery Foundation; Araclon Biotech; BioClinica, Inc.; Biogen; Bristol-Myers Squibb Company; CereSpir, Inc.; Cogstate; Eisai Inc.; Elan Pharmaceuticals, Inc.; Eli Lilly and Company; EuroImmun; F. Hoffmann-La Roche Ltd and its affiliated company Genentech, Inc.; Fujirebio; GE Healthcare; IXICO Ltd.; Janssen Alzheimer Immunotherapy Research & Development, LLC.; Johnson & Johnson Pharmaceutical Research & Development LLC.; Lumosity; Lundbeck; Merck & Co., Inc.; Meso Scale Diagnostics, LLC.; NeuroRx Research; Neurotrack Technologies; Novartis Pharmaceuticals Corporation; Pfizer Inc.; Piramal Imaging; Servier; Takeda Pharmaceutical Company; and Transition Therapeutics. The Canadian Institutes of Health Research is providing funds to support ADNI clinical sites in Canada. Private sector contributions are facilitated by the Foundation for the National Institutes of Health (www.fnih.org). The grantee organization is the Northern California Institute for Research and Education, and the study is coordinated by the Alzheimer's Therapeutic Research Institute at the University of Southern California. ADNI data are disseminated by the Laboratory for Neuro Imaging at the University of Southern California. The OASIS Cross-Sectional project (Principal Investigators: D. Marcus, R, Buckner, J, Csernansky J. Morris) was supported by the following grants: P50 AG05681, P01 AG03991, P01 AG026276, R01 AG021910, P20 MH071616, and U24 RR021382.

# Convolutional Neural Networks for Classification of Alzheimer's Disease: Overview and Reproducible Evaluation

## Supplementary Material

Junhao Wen[a,b,c,d,e*], Elina Thibeau-Sutre[a,b,c,d,e*], Mauricio Diaz-Melo[e,a,b,c,d], Jorge Samper-González[e,a,b,c,d], Alexandre Routier[e,a,b,c,d], Simona Bottani[e,a,b,c,d], Didier Dormont[e,a,b,c,d,f], Stanley Durrleman[e,a,b,c,d], Ninon Burgos[a,b,c,d,e], Olivier Colliot[a,b,c,d,e,f,g,†] , for the Alzheimer's Disease Neuroimaging Initiative and the Australian Imaging Biomarkers and Lifestyle flagship study of ageing

[a]*Institut du Cerveau et de la Moelle épinière, ICM, F-75013, Paris, France*

[b]*Sorbonne Université, F-75013, Paris, France*

[c]*Inserm, U 1127, F-75013, Paris, France*

[d]*CNRS, UMR 7225, F-75013, Paris, France*

[e]*Inria, Aramis project-team, F-75013, Paris, France*

[f]*AP-HP, Hôpital de la Pitié Salpêtrière, Department of Neuroradiology, F-75013, Paris, France*

[g]*AP-HP, Hôpital de la Pitié Salpêtrière, Department of Neurology, F-75013, Paris, France*

*denotes shared first authorship

We present additional methodological explanations, tables and figures in this supplementary material. More specifically, we first present in detail the methodology of our literature review (eMethod 1) and a brief introduction of the involved topics, namely deep learning, image preprocessing and AD (eMethod 2). We then describe the datasets used in our study in eMethod 3. From eTable1 to eTable3, we present the architecture hyperparameters for the chosen models. The training hyperparameters for autoencoder pre-training and classification are shown in eTable 4 and eTable 5, respectively. Lastly, the monitoring of training process, including the display of the training/validation loss and accuracy, is presented from eFigure 1 to eFigure 4.

**Table of contents:**

# eMethod 1. Literature search methodology

We searched PubMed and Scopus for articles published up to the time of the search (15th of January 2019). Our request contains words linked to four different concepts: Alzheimer's disease, classification, deep learning and neuroimaging. The words matching these concepts were identified in the abstracts and titles of the articles of a first bibliography done on Google Scholar. In Scopus a restriction was added to remove the articles linked to electroencephalography that appeared with our query and were out of our scope. This restriction was not applied in PubMed as it concerns only a few articles (less than 10). The line of the query linked to the neuroimaging concept was extended to all fields, as some authors do not mention at all in the title, abstract or keywords the modalities that they employed.

*Scopus query:*

TITLE-ABS-KEY ( alzheimer's  OR  alzheimer  OR  "Mild Cognitive Impairment" )
AND
TITLE-ABS-KEY ( classification  OR  diagnosis  OR  identification  OR  detection  OR  recognition )
AND
TITLE-ABS-KEY ( cnn  OR  "Convolutional Network"  OR  "Deep Learning"  OR  "Neural Network"  OR  autoencoder  OR  gan )
AND
ALL ( mri  OR  "Magnetic Resonance Imaging"  OR  "Structural Magnetic Resonance Imaging"  OR  neuroimaging  OR  brain-imaging )
AND NOT
TITLE-ABS-KEY ( eeg  OR  eegs  OR  electroencephalogram  OR  electroencephalographic )

*PubMed query:*

(alzheimer's [Title/Abstract] OR alzheimer [Title/Abstract] OR "Mild Cognitive Impairment" [Title/Abstract] )
AND
(cnn OR "Convolutional Network" [Title/Abstract] OR "Deep Learning" [Title/Abstract] OR "Neural Network" [Title/Abstract] OR autoencoder [Title/Abstract] OR gan [Title/Abstract] )
AND
(classification [Title/Abstract] OR diagnosis [Title/Abstract] OR identification [Title/Abstract] OR detection [Title/Abstract] OR recognition [Title/Abstract] )
AND
(mri OR "Magnetic Resonance Imaging" OR "Structural Magnetic Resonance Imaging" OR neuroimaging OR brain-imaging)

391 records were found with Scopus and 80 records were found with PubMed. After merging the two sets and removing duplicates, 406 records were identified. Before filtering the result, we removed from this list 10 conference proceedings books and 1 non-english article. We finally ended with 395 records to filter.

Once identified, all records were filtered in a 3-step process. We selected the records based on the abstract, the type and the content.

- **Record screening based on abstract**

During this step, the abstracts of the articles were read to keep only the methods corresponding to the following criteria:

- use of anatomical MRI (when the modality was specified),

- classification of AD stages, then we excluded papers using deep learning to preprocess, segment or complete data, as well as the classification of different diseases or classification of different symptoms in AD population (depression, ICD…),

- exclusion of animal models,

- exclusion of reviews.

We chose to exclude the 31 reviews of our set as none of them focused on our topic. We did not detail the reasons of the exclusion of the papers in the diagram as many papers cumulate several criteria of exclusion. After this screening phase, we were left with 124 records.

● **Record screening based on type**

Our search on PubMed and Scopus comprises only peer-reviewed items. However, there is a different level of peer-review between conference papers and journal articles, hence we kept all journal articles and recent conference papers (published since 2017). We decided to not only restrict to journal articles because it would have reduced the number of items to 48. We decided to keep recent conference papers because we considered that if the older ones were not transformed into journal articles it may mean that their contributions were not sufficient. After this step, the set contained 93 items.

● **Record screening based on content**

This step was mainly used to sort the papers between the different sections of our state-of-the art. We detected in this way papers that were out of the scope of our review (longitudinal and multimodal studies, deep learning techniques other than CNN). We excluded only 22 papers because of i) use of another modality (1 paper); ii) duplicate content (2 papers); iii) lack of explanation on the method employed (7 papers); iv) no access to the content (12 papers). This step was reviewed by another member of the team to confirm the exclusions. In the end, our search resulted in 71 conference and journal articles, including 32 that are centered on our topic.

Diagram summarizing the bibliographic methodology

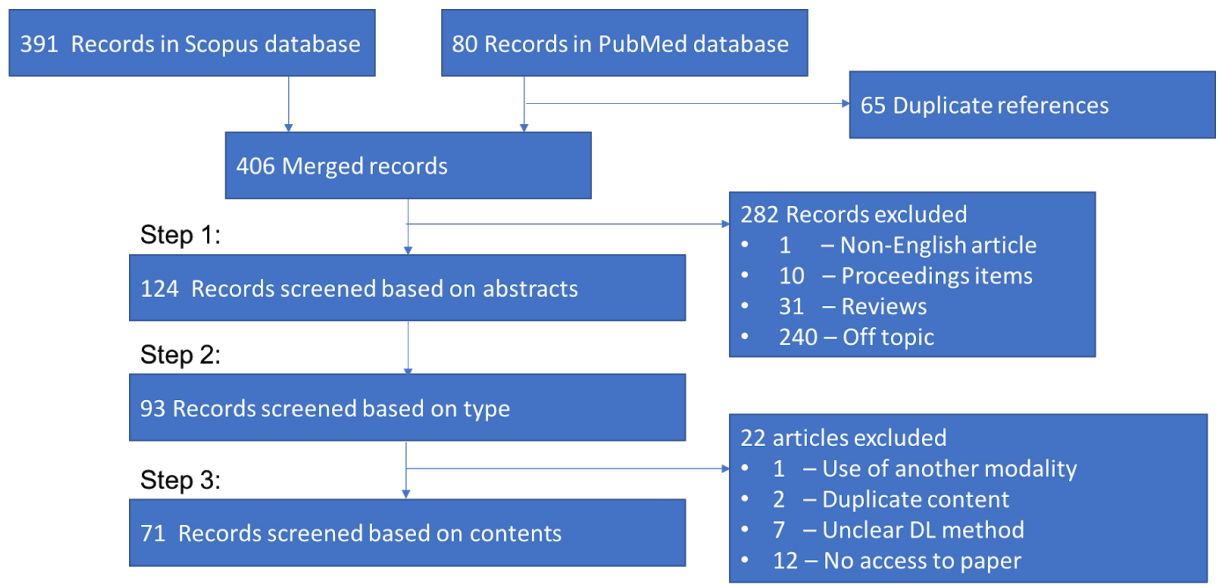

## eMethod 2. Convolutional neural networks

- **Main building layers of CNN**

CNNs are the most widely used type of network for computer vision and image analysis. A CNN is made of an input and an output layer, as well as different hidden layers. The hidden layers typically include convolutional layers, pooling layers, activation functions and fully connected (FC) layers.

The convolutional layer is the core building block of a CNN. It acts as an automatic feature extractor (on the contrary, conventional ML methods would typically use hand-craft feature extraction or selection). Convolutional layers apply learnable filters to all available receptive fields with a convolutional operation. A filter or kernel is a 2D (or 3D for MRI) matrix of weights. A receptive field is a local patch of the input image, of the same size as the filter. The filter is convolved with all the local receptive fields. The application of a given filter to the whole input image generates a feature map or activation map. All the feature maps are then stacked to constitute the output volume of a convolutional layer. Several hyperparameters (number of filters, stride size and padding size) control the size of the output volume (see (Dumoulin and Visin, 2016) for more details).

Another building block of CNNs is the pooling layer, which reduces the dimensionality of the feature maps. The pooling layer combines the outputs of a cluster of neurons of the current layer into a single neuron in the next layer (Ciresan et al., 2011; Krizhevsky et al., 2012). Pooling can be of different types, such as max, average and sum pooling (Scherer et al., 2010).

To learn a mapping between the adjacent convolutional layers, one applies activation functions to the output volume of each convolutional layer. The rectified linear unit (ReLU) is the most common activation function and ensures a sparse and non-linear representation (Glorot et al., 2011; Krizhevsky et al., 2012; Nair and Hinton, 2010). However, ReLU can be fragile during backpropagation. Indeed, the fact that ReLU sets all negative values to be zero can cause the problem of gradient vanishing or dying ReLU. If this happens, the gradient flowing through the unit will be forever zero during backpropagation. One alternative, leaky ReLU, can overcome this drawback by introducing a small negative slope (e.g. 0.01), thus allowing a small positive gradient when the unit is not active (Maas et al., 2013).

FC layers learn the relationship between the features, extracted by previous convolutional and pooling layers, and the target (in our case the patient's diagnosis). In a FC layer, all the neurons in the current layer are connected to all the neurons in the previous layer. The output volumes (one for each feature map) from the previous convolutional layers are first flattened and then fed as input to the FC. For a n-class classification problem, the output of the last FC layer is composed of n neurons which values indicate membership to a given class. This can be transformed into n probabilities by using a softmax function on the outputs (Goodfellow et al., 2016).

The loss function is used to measure the difference between the predicted and true labels. Cross entropy loss, measuring the distance between the output distribution and the real distribution, is widely used in classification tasks (de Boer et al., 2005). Other loss functions were also discussed in the literature, such as mean squared error (MSE) loss and hinge loss (see (Janocha and Czarnecki, 2017) for details).

The weights and biases of the network are learned using an optimization algorithm, such as the stochastic gradient descent (SGD). Most often, backpropagation is used to successively update the weights of the different layers.

- **Classical CNN architectures**

Several CNN architectures have become classical, often due to their performance on the ImageNet Large Scale Visual Recognition Challenge (ILSVRC): a benchmark in object category classification and detection on hundreds of object categories and millions of images (Deng et al., 2009). These architectures were originally designed for 2D natural images. However, some of them have been adapted to the applications of MRIs.

Before the ILSVRC that began in 2010, Yann Lecun proposed LeNet-5 to recognize handwritten digits from the MNIST database (Lecun et al., 1998). This network includes seven layers: two convolutional layers associated with pooling layers, followed by three FC layers.

In 2012, AlexNet (Krizhevsky et al., 2012) significantly outperformed all the prior competitors of the ILSVRC, reducing the top-5 error from 26% to 15.3%. The network went deeper than LeNet-5, with more filters per layer. It consisted of five convolutional layers with decreasing filter size (11x11, 5x5 and 3x3) and three FC layers.

The runner-up at ILSVRC 2014 was VGGNet (Simonyan and Zisserman, 2014), which consists of 16 convolutional layers. It was appealing because of its uniform architecture, including only 3x3 convolutional filters cross the entire architecture. One of the main conclusion of this architecture is that using many small filters of size 3x3 is more efficient than using only a few filters of bigger size. The winner of that year was GoogleNet or Inception V1 (Szegedy et al., 2015). It went deeper (22 layers) and achieved a top-5 error rate of 6.67%. This architecture was inspired by LeNet-5 and implemented a novel element called the inception layer. The idea behind the inception layer is to convolve over larger receptive fields, but also keep a fine resolution based on smaller receptive fields. Thus, different filter sizes (from 1x1 to 5x5) were used in the same convolutional layer.

ILSVRC 2015 was won by the Residual Neural Network (ResNet) (He et al., 2016) with a top-5 error rate of 3,57%. ResNet includes over a hundred layers by introducing a novel architecture with shortcut connections that perform identity mapping and heavy batch normalization. Such shortcut connections make the deep residual nets easier to optimize than their counterpart "plain" nets.

DenseNet was presented at ILSVRC 2016 (Huang et al., 2017). It introduces the so-called dense block: each layer receives the outputs of all previous layers as input. The underlying assumption of dense connectivity is that each layer should have access to all the preceding feature maps and this "collective knowledge" therefore helps to improve the performance. In the same way than ResNet, dense connectivity allows the construction of very deep CNNs, such as DenseNet-264 which consists of 264 layers.

- **Methods to deal with overfitting**

Neuroimaging datasets of AD patients are usually of relatively small size (typically a few hundreds of samples) compared, for instance, to those in computer vision (typically several million). DL models tend to easily overfit when trained on small samples due to the large number of learnt parameters (Goodfellow et al., 2016). Here, we summarize the main strategies to alleviate overfitting.

Data augmentation aims at generating new data samples from the available training data (Perez and Wang, 2017). It can be categorized into: i) transformation methods, which apply a combination of simple transformations (e.g. rotation, distortion, blurring and flipping) on the training data and ii) data synthesis methods, which aim to learn the training distribution to then generate new samples. Data synthesis often relies on autoencoders (AE) (Bourlard and Kamp, 1988; Hinton and Zemel, 1994; Yann, 1987) and Generative Adversarial Networks (GANs) (Goodfellow, 2016).

Dropout randomly and independently drops neurons, setting their output value to be zero along with their connections (Srivastava et al., 2014). This aims to make the network less complex and thus less prone to overfitting.

Another approach involves a regularization of the weights which makes the model less complex. This enhances the generalizability of the model. In DL, a common regularization is weight decay, where the updated weights are regularized by multiplying by a factor slightly smaller than 1 (Krogh and Hertz, 1992).

Batch normalization is a procedure which normalizes the input of a given set of layers (the normalization is done using the mean and standard-deviation of a batch, hence the name) (Ioffe and Szegedy, 2015). This procedure acts as a regularizer, in some cases eliminating the need for dropout (Ioffe and Szegedy, 2015). In addition, it helps battle against the gradient explosion phenomenon and allows using much higher learning rates and being less careful about initialization (Panigrahi et al., 2018).

Transfer learning is a broadly defined terminology. In general, it consists in using a model trained on a given task, called the source task (e.g. ImageNet classification task or unsupervised learning task), in order to perform a target task (e.g. AD classification). Here, we introduce two transfer learning approaches that have been used in the context of AD classification. The first one is based on performing unsupervised learning before the supervised learning on the task of interest. It is supposed to be useful when one has limited labeled data but a larger set of unlabeled data. In that case, the most common approach is to use an AE (Yann, 1987). Strictly speaking, the AE is made of two parts: an encoder layer and a decoder layer. Generally, several AEs are stacked, the resulting being called stacked AE, but which we will refer to as AE for the sake of simplicity. The encoder learns to compress the original data and produces a representation, the decoder then reconstructs the input using only this representation. An illustration of AE is shown in Figure 1. The weights and biases of the target network (e.g. CNNs) are then initialized with those of the encoder part of AE, which should provide a better initialization. The second approach involves transferring a model trained on ImageNet to the problem of classification of AD. As for the AE, the weights and biases of the target network are initialized with those of the source network. The idea behind is that random weight initialization of DL models may place parameters in a region of the parameter space where poor generalization occurs, while transfer learning may provide a better initialization (Erhan et al., 2010).

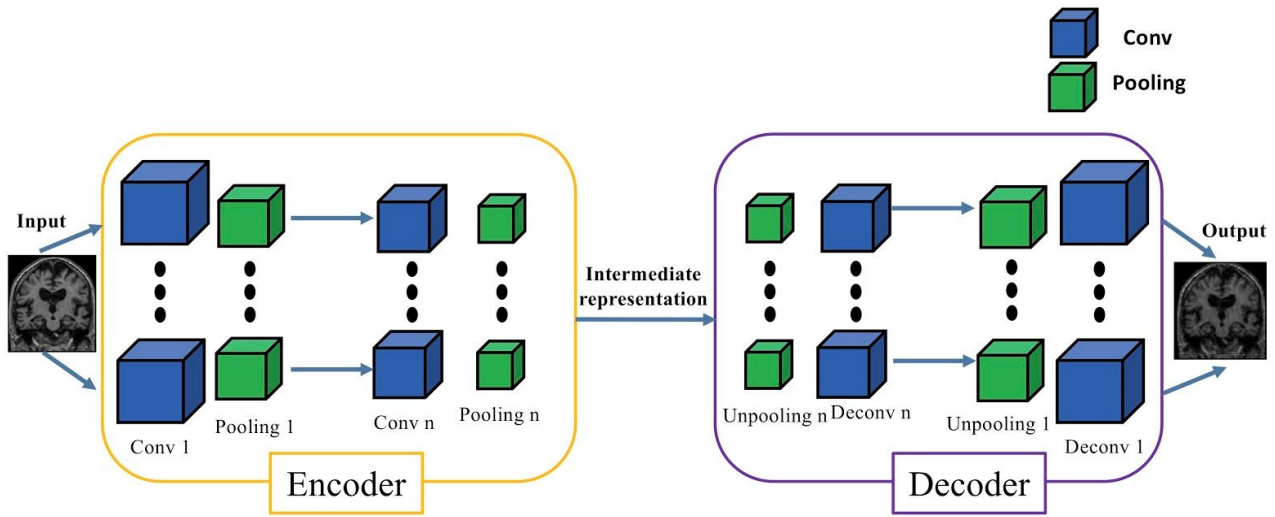

**Figure 1**. The architecture of a stacked AE made up of n AEs. For the n-th AE, the encoder part is made of one convolutional layer (Conv n), one pooling layer (Pooling n) and one activation function, sequentially. Correspondingly, the decoder part consists of one activation function, one unpooling layer (Unpooling n) and one deconvolutional layer (Deconv n), sequentially. The n AEs's encoder and decoder parts are separated and then stacked to construct the final stacked AE. This layer-wise fashion ensures that the output of the former AE's encoder/decoder connects as input for the next AE's encoder/decoder. The intermediate representation provides a reduced representation of the data.

Early Stopping consists in stopping the learning process at an earlier point. It aims to determine the number of epochs (or iterations) at which the network should be stopped before being severely overfitted. For instance, one can select the model parameters corresponding to the lowest validation error rather than the last updated parameters. Various other stopping criterions have been proposed (Prechelt, 2012; Yao et al., 2007; Zhang and Yu, 2005).

## eMethod 3. MRI preprocessing

A proper image preprocessing procedure is a fundamental step to ensure a good classification performance, especially in the domain of MRI (Cuingnet et al., 2011; Lu and Weng, 2007; Uchida, 2013). Although CNNs have the potential to extract low-to-high level features from the raw images, the influence of image preprocessing remains to be clarified. We present here the essential steps for MR image processing in the context of AD classification.

- **Bias field correction**

MR images can be corrupted by a low frequency and smooth signal caused by magnetic field inhomogeneities. This bias field induces variations in the intensity of the same tissue in different locations of the image, which deteriorates the performance of image analysis algorithms such as registration (Vovk et al., 2007). Several methods exist to correct these intensity inhomogeneities, two popular ones being the nonparametric nonuniformity intensity normalization (N3) algorithm (Sled et al., 1998), available for example in the Freesurfer software package[1], and the N4 algorithm (Tustison et al., 2010) implemented in ITK[2].

- **Intensity rescaling and standardization**

As MRI is usually not a quantitative imaging modality, MR images usually have different intensity ranges and the intensity distribution of the same tissue type may be different between two images, which might affect the subsequent image preprocessing steps. The first point can be dealt with by globally rescaling the image, for example between 0 and 1 using the minimum and maximum intensity values (Juszczak et al., 2002). Intensity standardization can be achieved using techniques such as histogram matching (Madabhushi and Udupa, 2005).

- **Skull stripping**

Extracranial tissues can be an obstacle for image analysis algorithms (Kalavathi and Prasath, 2016). A large number of methods have been developed for brain extraction, also called skull stripping, and many are implemented in software tools, such as the Brain Extraction Tool (BET) (Smith, 2002) available in FSL[3], or the Brain Surface Extractor (BSE) (Shattuck et al., 2001) available in BrainSuite[4]. These methods are often sensitive to the presence of noise and artifacts, which can result in over or under segmentation of the brain.

- **Image registration**

Medical image registration consists of spatially aligning two or more images, either globally (rigid and affine registration) or locally (non-rigid registration), so that voxels in corresponding positions contain comparable information. A large number of software tools have been developed for MRI-based registration (Oliveira and Tavares, 2014). FLIRT[5] (Greve and Fischl, 2009; Jenkinson et al., 2002; Jenkinson and Smith, 2001) and FNIRT[6] (Andersson et al., 2010) are FSL tools dedicated to linear and non-linear registration, respectively. The Statistical Parametric Mapping (SPM) software package[7] and Advanced Normalization Tools[8] (ANTs) also offer solutions for both linear and non-linear registration (Ashburner and Friston, 2000; Avants et al., 2014; Friston et al., 1995).

---

[1] http://surfer.nmr.mgh.harvard.edu/fswiki/recon-all
[2] http://hdl.handle.net/10380/3053
[3] https://fsl.fmrib.ox.ac.uk/fsl/fslwiki/BET/UserGuide
[4] http://brainsuite.org/processing/surfaceextraction/bse
[5] https://fsl.fmrib.ox.ac.uk/fsl/fslwiki/FLIRT
[6] https://fsl.fmrib.ox.ac.uk/fsl/fslwiki/FNIRT
[7] https://www.fil.ion.ucl.ac.uk/spm
[8] http://stnava.github.io/ANTs

# eMethod 4. Main datasets used for AD classification

Three publicly available datasets have been mainly used for the study of AD: the Alzheimer's Disease Neuroimaging Initiative (ADNI), the Australian Imaging, Biomarkers and Lifestyle (AIBL) and the Open Access Series of Imaging Studies (OASIS). In the following, we briefly describe these datasets and provide explanations on the diagnosis labels provided. Indeed, the diagnostic criteria of these studies differ, hence there is no strict equivalence between the labels of ADNI and AIBL, and those of OASIS.

- **Alzheimer's Disease Neuroimaging Initiative (ADNI)**

Part of the data used in the preparation of this article were obtained from the Alzheimer's Disease Neuroimaging Initiative (ADNI) database (adni.loni.usc.edu). The ADNI was launched in 2003 as a public-private partnership, led by Principal Investigator Michael W. Weiner, MD. The primary goal of ADNI has been to test whether serial magnetic resonance imaging (MRI), positron emission tomography (PET), other biological markers, and clinical and neuropsychological assessment can be combined to measure the progression of mild cognitive impairment (MCI) and early Alzheimer's disease (AD). For up-to-date information, see www.adni-info.org.1 The ADNI study is composed of 4 cohorts: ADNI-1, ADNI-GO, ADNI-2 and ADNI-3. These cohorts are dependant and longitudinal, meaning that one cohort may include the same patient more than once and that different cohorts may include the same patients. Diagnosis labels are given by a physician after a series of tests (Petersen et al., 2010). The existing labels are:

- AD (Alzheimer's disease): mildly demented patients,
- MCI (mild cognitive impairment): patients in the prodromal phase of AD,
- NC (normal controls): elderly control participants,
- SMC (significant memory concern): participants with cognitive complaints and no pathological neuropsychological findings. The designations SMC and subjective cognitive decline (SCD) are equivalently found in the literature.

Since the ADNI-GO and ADNI-2 cohorts, new patients at the very beginning of the prodromal stage have been recruited (Aisen et al., 2010), hence the MCI label has been split into two labels:

- EMCI (early MCI): patients at the beginning of the prodromal phase,
- LMCI (late MCI): patients at the end of the prodromal phase (similar to the previous label MCI of ADNI-1).

This division is made on the basis of the score obtained on memory tasks corrected by the education level. However, both classes remain very similar and they are fused in many studies under the MCI label.

- **Australian Imaging, Biomarkers and Lifestyle (AIBL)**

We also used data collected by the AIBL study group. Similarly to ADNI, the Australian Imaging, Biomarker & Lifestyle Flagship Study of Ageing seeks to discover which biomarkers, cognitive characteristics, and health and lifestyle factors determine the development of AD. The AIBL project includes a longitudinal cohort of patients. Several modalities are present in the dataset, such as clinical and imaging (MRI and PET) data, as well as the analysis of blood and CSF samples. As in ADNI, the diagnosis is given according to a series of clinical tests (Ellis et al., 2010, 2009) and the existing labels are AD, MCI and NC.

- **Open Access Series of Imaging Studies (OASIS)**

Finally, we used data from OASIS (http://www.oasis-brains.org/). This project includes three cohorts, OASIS-1, OASIS-2 and OASIS-3. The first cohort is only cross-sectional, whereas the other two are longitudinal. Available data is far more limited than in ADNI with only few clinical tests and imaging data (MRI and PET only in OASIS-3). Diagnosis labels are given only based on the clinical dementia rating (CDR) scale (Marcus et al., 2007). Two labels can be found in the OASIS-1 dataset:

- AD, which corresponds to patients with a non-null CDR score. This class gathers patients who would be spread between the MCI and AD classes in ADNI. A subdivision of this class is done based on the CDR, the scores of 0.5, 1, 2 and 3 representing very mild, mild, moderate and severe dementia, respectively.
- Control, which corresponds to patients with a CDR of zero. Unlike ADNI, some of the controls are younger than 55.

## eTable 1. Architecture hyperparameters for 3D subject-level CNN

As the architecture depends on the size of the input, it slightly differs between the two types of preprocessing (i.e. "Minimal" or "Extensive"). This difference only affects the size of the input of the first FC layer (FC1). The output size of each layer is reported depending on the preprocessing used in the last two columns.

The padding size in convolutional layers has been set to 1 not to decrease the size of the convolutional layer outputs. Without any padding, the number of nodes at the end of the last convolutional layer is too small to reconstruct the image correctly using an autoencoder for the Extensive preprocessing.

The padding size in pooling layers depends on the input: columns of zeros are added along a dimension until the size along this dimension is a multiple of the stride size.

| Layer | Filter size | Number of filters / neurons | Stride size | Padding size | Dropout rate | Output size (Minimal) | Output size (Extensive) |
|---|---|---|---|---|---|---|---|
| Conv1+BN+ReLU | 3x3x3 | 8 | 1 | 1 | -- | 8x169x208x179 | 8x121x145x121 |
| MaxPool1 | 2x2x2 | -- | 2 | adaptive | -- | 8x85x104x90 | 8x61x73x61 |
| Conv2+BN+ReLU | 3x3x3 | 16 | 1 | 1 | -- | 16x85x104x90 | 16x61x73x61 |
| MaxPool2 | 2x2x2 | -- | 2 | adaptive | -- | 16x43x52x45 | 16x31x37x31 |
| Conv3+BN+ReLU | 3x3x3 | 32 | 1 | 1 | -- | 32x43x52x45 | 32x31x37x31 |
| MaxPool3 | 2x2x2 | -- | 2 | adaptive | -- | 32x22x26x23 | 32x16x19x16 |
| Conv4+BN+ReLU | 3x3x3 | 64 | 1 | 1 | -- | 64x22x26x23 | 64x16x19x16 |
| MaxPool4 | 2x2x2 | -- | 2 | adaptive | -- | 64x11x13x12 | 64x8x10x8 |
| Conv5+BN+ReLU | 3x3x3 | 128 | 1 | 1 | -- | 128x11x13x12 | 128x8x10x8 |
| MaxPool5 | 2x2x2 | -- | 2 | adaptive | -- | 128x6x7x6 | 128x4x5x4 |
| Dropout | -- | -- | -- | -- | 0.5 | 128x6x7x6 | 128x4x5x4 |
| FC1 | -- | 1300 | -- | -- | -- | 1300 | 1300 |
| FC2 | -- | 50 | -- | -- | -- | 50 | 50 |
| FC3 | -- | 2 | -- | -- | -- | 2 | 2 |
| Softmax | -- | -- | -- | -- | -- | -- | 2 |

BN: batch normalization; Conv: convolutional layer; FC: fully connected; MaxPool: max pooling.

**eTable 2. Architecture hyperparameters for 3D ROI-based and patch-level CNN**

The padding size in pooling layers depends on the input: columns of zeros are added along a dimension until the size along this dimension is a multiple of the stride size.

| Layer | Filter size | Number of filters / neurons | Stride size | Padding size | Output size |
|---|---|---|---|---|---|
| Conv1+BN+ReLU | 3x3x3 | 15 | 1 | 0 | 15x48x48x48 |
| MaxPool1 | 2x2x2 | -- | 2 | adaptive | 15x24x24x24 |
| Conv2+BN+ReLU | 3x3x3 | 25 | 1 | 0 | 25x22x22x22 |
| MaxPool2 | 2x2x2 | -- | 2 | adaptive | 25x11x11x11 |
| Conv3+BN+ReLU | 3x3x3 | 50 | 1 | 0 | 50x9x9x9 |
| MaxPool3 | 2x2x2 | -- | 2 | adaptive | 50x5x5x5 |
| Conv4+BN+ReLU | 3x3x3 | 50 | 1 | 0 | 50x3x3x3 |
| MaxPool4 | 2x2x2 | -- | 2 | adaptive | 50x2x2x2 |
| FC1 | -- | 50 | -- | -- | 50 |
| FC2 | -- | 40 | -- | -- | 40 |
| FC3 | -- | 2 | -- | -- | 2 |
| Softmax | -- | -- | -- | -- | 2 |

BN: batch normalization; Conv: convolutional layer; FC: fully connected; MaxPool: max pooling.

**eTable 3. Architecture hyperparameters for 2D slice-level CNN**

Table B explicits the architecture of our 2D slice-level CNN. Shortcuts are displayed with arrows and are adding the two feature maps linked together and applying ReLU to form a new feature map given to the following layer.

When shortcuts are linking feature maps of different sizes, the arrow is associated with a downsampling layer (see table A) applied to the largest feature map.

A. Characteristics of the downsampling layers

| Layer | Filter size | Number of filters / neurons | Stride size | Padding size | Dropout rate |
|---|---|---|---|---|---|
| Conv8 | 1x1 | 128 | 2 | 0 | -- |
| Conv13 | 1x1 | 256 | 2 | 0 | -- |
| Conv18 | 1x1 | 512 | 2 | 0 | -- |

B. Architecture of the 2D slice-level CNN (adaptation of the ResNet-18)

| Layer | Filter size | Number of filters / neurons | Stride size | Padding size | Dropout rate | Output size |
|---|---|---|---|---|---|---|
| Conv1+BN+ReLU | 7x7 | 64 | 2 | 3 | -- | 64x112x112 |
| MaxPool1 | 3x3 | -- | 2 | 1 | -- | 64x56x56 |
| Conv2+BN+ReLU | 3x3 | 64 | 1 | 1 | -- | 64x56x56 |
| Conv3+BN | 3x3 | 64 | 1 | 1 | -- | 64x56x56 |
| Conv4+BN+ReLU | 3x3 | 64 | 1 | 1 | -- | 64x56x56 |
| Conv5+BN | 3x3 | 64 | 1 | 1 | -- | 64x56x56 |
| Conv6+BN+ReLU | 3x3 | 128 | 2 | 1 | -- | 128x28x28 |
| Conv7+BN | 3x3 | 128 | 1 | 1 | -- | 128x28x28 |
| Conv9+BN+ReLU | 3x3 | 128 | 1 | 1 | -- | 128x28x28 |
| Conv10+BN | 3x3 | 128 | 1 | 1 | -- | 128x28x28 |
| Conv11+BN+ReLU | 3x3 | 256 | 2 | 1 | -- | 256x14x14 |
| Conv12+BN | 3x3 | 256 | 1 | 1 | -- | 256x14x14 |
| Conv14+BN+ReLU | 3x3 | 256 | 1 | 1 | -- | 256x14x14 |
| Conv15+BN | 3x3 | 256 | 1 | 1 | -- | 256x14x14 |
| Conv16+BN+ReLU | 3x3 | 512 | 2 | 1 | -- | 512x7x7 |
| Conv17+BN | 3x3 | 512 | 1 | 1 | -- | 512x7x7 |
| Conv19+BN+ReLU | 3x3 | 512 | 1 | 1 | -- | 512x7x7 |
| Conv20+BN | 3x3 | 512 | 1 | 1 | -- | 512x7x7 |
| AveragePool1 | 7x7 | -- | 1 | 0 | -- | 512x1x1 |
| FC1 | -- | 1000 | -- | -- | -- | 1000 |
| Dropout | -- | -- | -- | -- | 0.8 | 1000 |
| FC2 | -- | 2 | -- | -- | -- | 2 |
| Softmax | -- | -- | -- | -- | -- | 2 |

# eTable 4. Training hyperparameters for classification experiments

A summary of the experiments can be found in Table A. The corresponding hyperparameters are listed in Table B indicated by the experiments numbers.

Common hyperparameters for all experiments: optimizer: Adam; Adam parameters: betas=(0.9, 0.999), epsilon=1e-8; loss: cross entropy.

When transfer learning is applied, the corresponding experiment number is given between brackets and can be found in eTable 5 for AE pretraining (AE) and eTable 4 for cross-task transfer learning (CTT).

### A. Summary of experiments performed

| Experiment number | Classification architectures | Training data | Image preprocessing | Intensity rescaling | Data split | Training approach | Transfer learning | Task |
|---|---|---|---|---|---|---|---|---|
| 1 | 3D subject-level CNN | Baseline | Minimal | None | subject-level | single-CNN | None | AD vs CN |
| 2 | | | | MinMax | | | | |
| 3 | | | | | | | AE (1) | |
| 4 | | Longitudinal | Minimal | MinMax | subject-level | single-CNN | AE (1) | |
| 5 | | | Extensive | | | | AE (2) | |
| 6 | | | Minimal | MinMax | subject-level | single-CNN | CTT (4) | sMCI vs pMCI |
| 7 | | Baseline | | | | | CTT (3) | |
| 8 | 3D ROI-based CNN | Baseline | Minimal | MinMax | subject-level | single-CNN | AE (3) | AD vs CN |
| 9 | | | | | | | CTT (8) | sMCI vs pMCI |
| 10 | | Longitudinal | | | | | AE (4) | AD vs CN |
| 11 | | | | | | | CTT (10) | sMCI vs pMCI |
| 12 | 3D patch-level CNN | Baseline | Minimal | MinMax | subject-level | single-CNN | AE (5) | AD vs CN |
| 13 | | Longitudinal | | | | | AE (6) | |
| 14 | | Baseline | Minimal | MinMax | subject-level | multi-CNN | AE (7) | AD vs CN |
| 15 | | | | | | | CTT (14) | sMCI vs pMCI |
| 16 | | Longitudinal | | | | | AE (8) | AD vs CN |
| 17 | | | | | | | CTT (16) | sMCI vs pMCI |
| 18 | 2D slice-level CNN | Baseline | Minimal | MinMax | subject-level | single-CNN | ImageNet pre-train | AD vs CN |
| 19 | | Longitudinal | | | | | | |
| 20 | | Baseline | Minimal | MinMax | slice-level (**data leakage**) | single-CNN | ImageNet pre-train | AD vs CN |

B. Hyperparameters corresponding to experiments described in Table A.

| Approach | Experiment | Number of epochs | Learning rate | Batch size | Dropout rate | Weight decay | Patience |
|----------|-----------|------------------|---------------|------------|--------------|--------------|----------|
| 3D subject-level CNN | 1 | 50 | 1e-4 | 12 | 0.5 | 0 | 10 |
| | 2 | 50 | 1e-4 | 12 | 0.5 | 0 | 10 |
| | 3 | 50 | 1e-4 | 12 | 0.5 | 0 | 10 |
| | 4 | 50 | 1e-4 | 12 | 0.5 | 0 | 5 |
| | 5 | 50 | 1e-4 | 12 | 0.5 | 0 | 5 |
| | 6 | 50 | 1e-5 | 12 | 0.5 | 0 | 10 |
| | 7 | 50 | 1e-5 | 12 | 0.5 | 0 | 20 |
| 3D ROI-based CNN | 8 | 200 | 1e-5 | 32 | -- | 1e-4 | 10 |
| | 9 | 200 | 1e-5 | 32 | -- | 1e-3 | 20 |
| | 10 | 200 | 1e-5 | 32 | -- | 1e-4 | 10 |
| | 11 | 200 | 1e-5 | 32 | -- | 1e-3 | 20 |
| 3D patch-level CNN | 12 | 200 | 1e-5 | 32 | -- | 1e-3 | 20 |
| | 13 | 200 | 1e-5 | 32 | -- | 1e-3 | 20 |
| | 14 | 200 | 1e-5 | 32 | -- | 1e-4 | 15 |
| | 15 | 200 | 1e-5 | 32 | -- | 1e-3 | 20 |
| | 16 | 200 | 1e-5 | 32 | -- | 1e-4 | 15 |
| | 17 | 200 | 1e-5 | 32 | -- | 1e-3 | 20 |
| 2D slice-level CNN | 18 | 50 | 1e-6 | 32 | 0.8 | 1e-4 | 15 |
| | 19 | 100 | 1e-6 | 32 | 0.8 | 1e-4 | 15 |
| | 20 | 50 | 1e-6 | 32 | 0.8 | 1e-4 | 15 |

**eTable 5. Training hyperparameters for autoencoder pretraining experiments**

A summary of the experiments can be found in table A. The corresponding hyperparameters are listed in Table B using the same experiments numbers.

Common hyperparameters for all experiments: optimizer: Adam; Adam parameters: betas=(0.9, 0.999), epsilon=1e-8; loss: mean squared entropy loss; training data: AD + MCI + CN; data split: subject-level. The stopping criterion is the maximal number of epochs.

A. Summary of autoencoder pretraining experiments performed.

| Experiment number | Classification architectures | Training data | Image preprocessing | Intensity rescaling | Training approach |
|---|---|---|---|---|---|
| 1 | 3D subject-level CNN | Baseline | Minimal | MinMax | single-CNN |
| 2 | | | Extensive | | |
| 3 | 3D ROI-based CNN | Baseline | Minimal | MinMax | single-CNN |
| 4 | | Longitudinal | | | |
| 5 | 3D patch-level CNN | Baseline | Minimal | MinMax | single-CNN |
| 6 | | Longitudinal | | | |
| 7 | | Baseline | | | multi-CNN |
| 8 | | Longitudinal | | | |

B. Hyperparameters corresponding to autoencoder pretraining experiments described in Table A.

| Approach | Experiment | Number of epochs | Learning rate | Batch size | Weight decay |
|---|---|---|---|---|---|
| 3D subject-level CNN | 1 | 50 | 1e-4 | 12 | 0 |
| | 2 | 30 | 1e-4 | 12 | 0 |
| 3D ROI-based CNN | 3 | 200 | 1e-5 | 32 | 0 |
| | 4 | 100 | 1e-5 | 32 | 0 |
| 3D patch-level CNN | 5 | 20 | 1e-5 | 32 | 0 |
| | 6 | 15 | 1e-5 | 32 | 0 |
| | 7 | 20 | 1e-5 | 32 | 0 |
| | 8 | 15 | 1e-5 | 32 | 0 |

**eFigure 1. Training process monitoring for 3D subject-level CNN**

Training and validation accuracy/loss during the training process were evaluated after the forward pass of 20 batches. The accuracy and loss curves were smoothed with a threshold (0.6).

For each plot a subfigure letter is used and the corresponding information on the experiment may be found in the table below or in eTable 4 according to the "Experiment number".

| Subfigure | Experiment number | Fold displayed | Epoch where training stopped | Epoch of the highest validation accuracy | Highest validation accuracy |
|-----------|-------------------|----------------|------------------------------|------------------------------------------|-----------------------------|
| A | 1 | 2 | 32 | 22 | 0.50 |
| B | 2 | 2 | 37 | 27 | 0.87 |
| C | 3 | 2 | 28 | 28 | 0.83 |
| D | 4 | 2 | 11 | 10 | 0.87 |
| E | 5 | 2 | 10 | 9 | 0.91 |
| F | 6 | 2 | 23 | 1 | 0.77 |
| G | 7 | 2 | 42 | 37 | 0.77 |

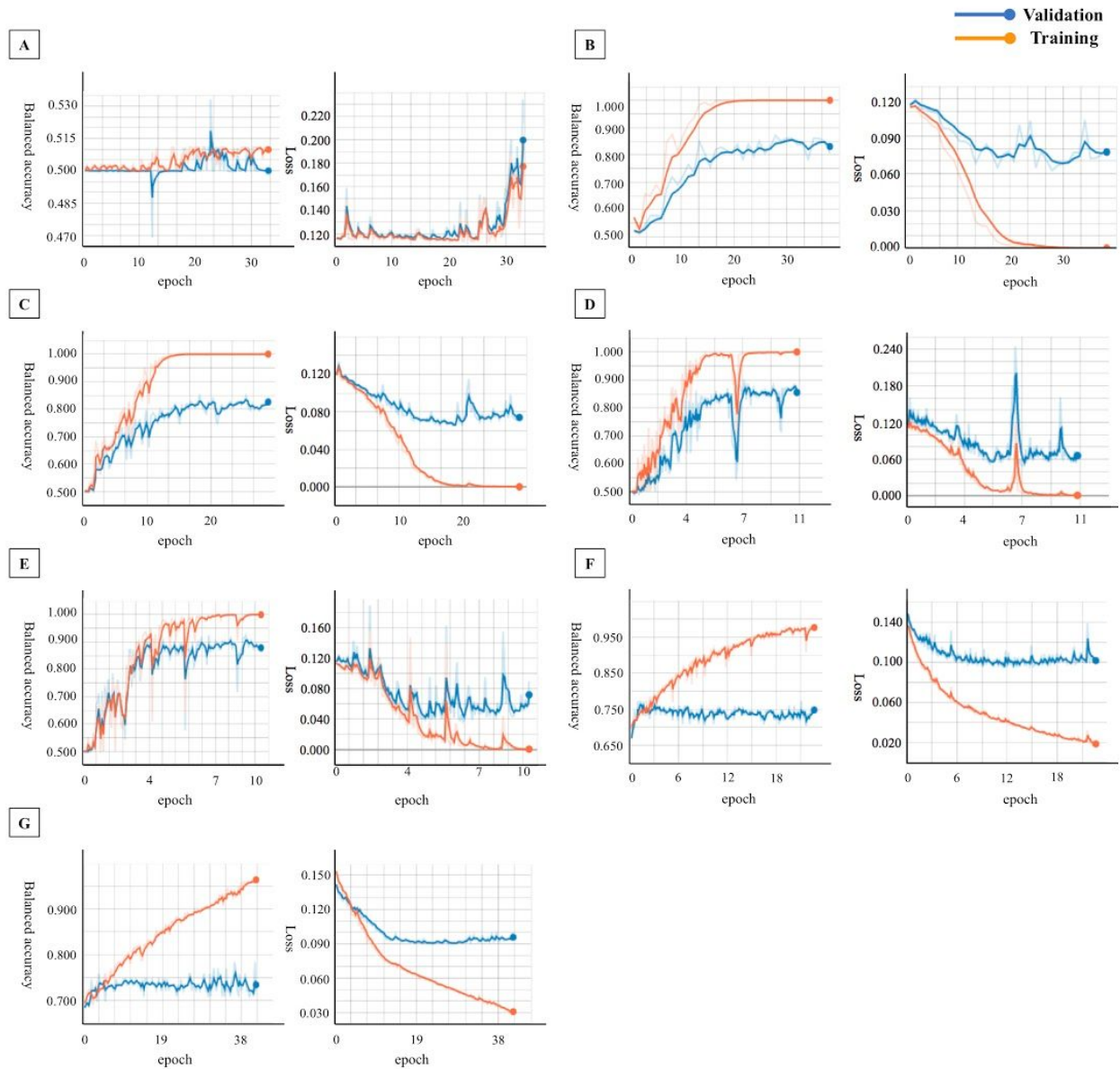

## eFigure 2. Training process monitoring for 3D ROI-based CNN

Training and validation accuracy/loss during the training process were evaluated after each epoch. The accuracy and loss curves were smoothed with a threshold (0.6).

For each plot a subfigure letter is used and the corresponding information on the experiment may be found in the table below or in eTable 4 according to the "Experiment number".

| Subfigure | Experiment number | Fold displayed | Epoch where training stopped | Epoch of the highest validation accuracy | Highest validation accuracy |
|-----------|-------------------|----------------|------------------------------|------------------------------------------|----------------------------|
| A | 8 | 2 | 141 | 125 | 0.89 |
| B | 9 | 1 | 60 | 60 | 0.84 |
| C | 10 | 3 | 89 | 51 | 0.88 |
| D | 11 | 3 | 55 | 48 | 0.82 |

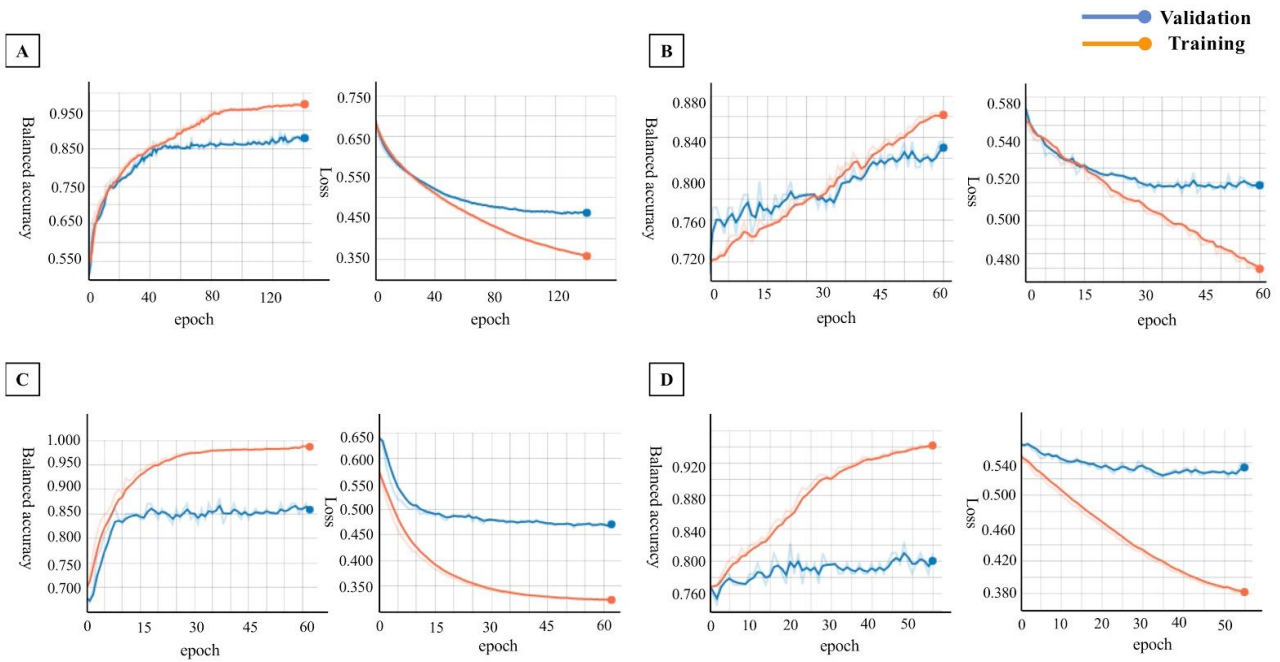

## eFigure 3. Training process monitoring for 3D patch-level CNN

Training and validation accuracy/loss during the training process were evaluated after each epoch. The accuracy and loss curves were smoothed with a threshold (0.6).

For each plot a subfigure letter is used and the corresponding information on the experiment may be found in the table below or in eTable 4 according to the "Experiment number". For multi-CNN experiments the CNN number is provided.

| Subfigure | Experiment number | Fold displayed | CNN number | Epoch where training stopped | Epoch of the highest validation accuracy | Highest validation accuracy |
|---|---|---|---|---|---|---|
| A | 12 | 2 | -- | 51 | 31 | 0.83 |
| B | 13 | 1 | -- | 44 | 42 | 0.77 |
| C | 14 | 1 | 5 | 110 | 110 | 0.85 |
| D | 15 | 1 | 19 | 148 | 46 | 0.80 |
| E | 16 | 1 | 29 | 96 | 86 | 0.81 |
| F | 17 | 1 | 19 | 58 | 30 | 0.78 |

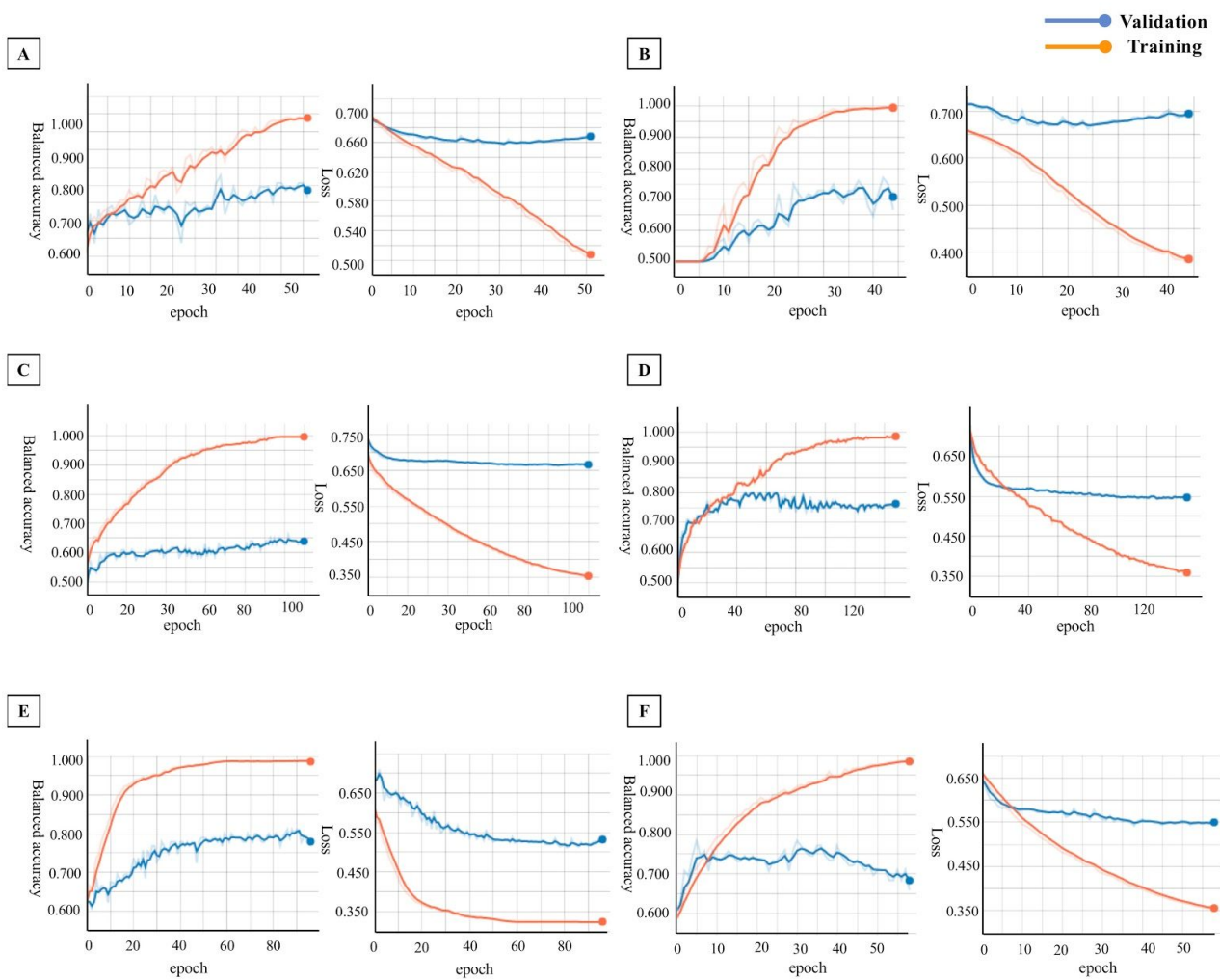

**eFigure 4. Training process monitoring for 2D slice-level CNN**

Training and validation accuracy/loss during the training process were evaluated after each epoch. The accuracy and loss curves were smoothed with a threshold (0.6).

For each plot a subfigure letter is used and the corresponding information on the experiment may be found in the table below or in eTable 4 according to the "Experiment number".

| Subfigure | Experiment number | Fold displayed | Epoch where training stopped | Epoch of the highest validation accuracy | Highest validation accuracy |
|-----------|-------------------|----------------|------------------------------|------------------------------------------|-----------------------------|
| A | 18 | 2 | 21 | 11 | 0.85 |
| B | 19 | 2 | 15 | 15 | 0.84 |
| C | 20 | 2 | 49 | 49 | 1.00 |

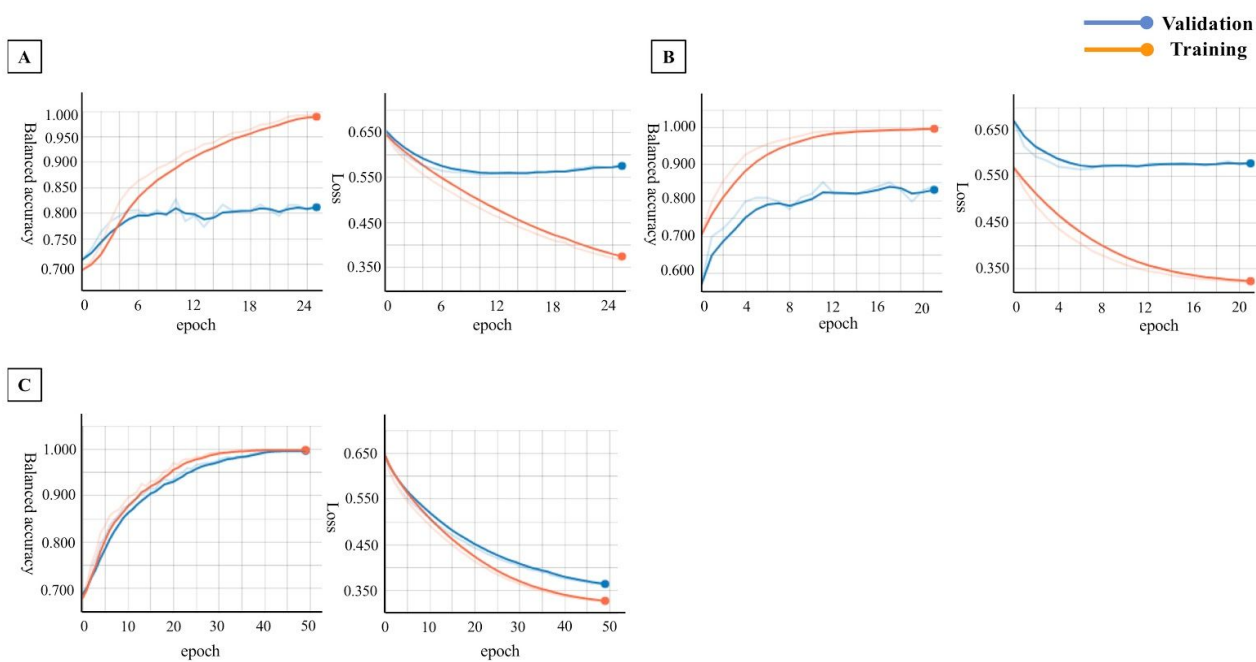

**eTable 6. Experiments performed with the single-CNN using thresholding**

MinMax: for CNNs, intensity rescaling was done based on min and max values, resulting in all values to be in the range of [0, 1]; AE: autoencoder.

The less informative patches (balanced accuracy < 0.7) were not included in the soft voting with the consideration that the labels' probabilities of these patches could harm the majority voting system.

| Classification architectures | Training data | Image preprocessing | Intensity rescaling | Data split | Training approach | Transfer learning | Task | Validation balanced accuracy |
|---|---|---|---|---|---|---|---|---|
| 3D patch-level CNN | Baseline | Minimal | MinMax | subject-level | single-CNN | AE pre-train | AD vs CN | 0.79 ± 0.03 [0.81, 0.81, 0.82, 0.80, 0.72] |
| | Longitudinal | | | | | | | 0.83 ± 0.03 [0.86, 0.85, 0.82, 0.84, 0.77] |

**eTable 7. Experiments performed with the multi-CNN without thresholding**

MinMax: for CNNs, intensity rescaling was done based on min and max values, resulting in all values to be in the range of [0, 1]; AE: autoencoder. sMCI vs pMCI tasks were done with as follows: the weights and biases of the model learnt on the source task (AD vs CN) were transferred to a new model fine-tuned on the target task (sMCI vs pMCI).

All the classifiers composing the multi-CNN were included in the soft voting.

| Classification architectures | Training data | Image preprocessing | Intensity rescaling | Data split | Training approach | Transfer learning | Task | Validation balanced accuracy |
|---|---|---|---|---|---|---|---|---|
| 3D patch-level CNN | Baseline | Minimal | MinMax | subject-level | multi-CNN | AE pre-train | AD vs CN | 0.76 ± 0.05 [0.74, 0.85, 0.73, 0.77, 0.69] |
| | | | | | | | sMCI vs pMCI | 0.72 ± 0.07 [0.78, 0.65, 0.61, 0.78, 0.76] |
| | Longitudinal | | | | | | AD vs CN | 0.72 ± 0.04 [0.74, 0.78, 0.72, 0.69, 0.66] |
| | | | | | | | sMCI vs pMCI | 0.70 ± 0.07 [0.73, 0.66, 0.61, 0.81, 0.69] |